\documentclass[nohyperref]{article}

\usepackage{microtype}
\usepackage{graphicx}
\usepackage{booktabs} % for professional tables

\usepackage{float}
\usepackage{amsmath, amssymb}
\usepackage{subfig}
\usepackage{centernot}
\usepackage{hyperref}

\usepackage[accepted]{icml2022}

\usepackage{amsmath}
\usepackage{amssymb}
\usepackage{mathtools}
\usepackage{amsthm}

\usepackage[capitalize,noabbrev]{cleveref}

\theoremstyle{plain}
\newtheorem{theorem}{Theorem}[section]

\theoremstyle{definition}
\newtheorem{definition}[theorem]{Definition}

\theoremstyle{remark}
\newtheorem{remark}[theorem]{Remark}

\usepackage[textsize=tiny]{todonotes}

\icmltitlerunning{Equal Experience in Recommender Systems}

\usepackage{environ}         % provides \BODY
\usepackage{etoolbox}        % provides \ifdimcomp
\usepackage{graphicx}        % provides \resizebox

\newlength{\myl}
\let\origequation=\equation
\let\origendequation=\endequation

\RenewEnviron{equation}{
  \settowidth{\myl}{$\BODY$}                       % calculate width and save as \myl
  \origequation
  \ifdimcomp{\the\linewidth}{>}{\the\myl}
  {\ensuremath{\BODY}}                             % True
  {\resizebox{\linewidth}{!}{\ensuremath{\BODY}}}  % False
  \origendequation
}

\begin{document}

\twocolumn[
\icmltitle{Equal Experience in Recommender Systems}



\icmlsetsymbol{equal}{*}

\begin{icmlauthorlist}
\icmlauthor{Jaewoong Cho}{comp,equal}
\icmlauthor{Moonseok Choi}{sch}
\icmlauthor{Changho Suh}{sch}
% % \icmlauthor{Firstname7 Lastname7}{comp}
% \icmlauthor{}{sch}
\end{icmlauthorlist}

\icmlaffiliation{comp}{KRAFTON Inc., Seoul, Korea}
\icmlaffiliation{sch}{KAIST, Daejeon, Korea}

\icmlcorrespondingauthor{Changho Suh}{chsuh@kaist.ac.kr}

\icmlkeywords{Machine Learning, ICML}

\vskip 0.3in
]



\printAffiliationsAndNotice{}  % leave blank if no need to mention equal contribution

\begin{abstract}
We explore the \emph{fairness} issue that arises in recommender systems. Biased data due to inherent stereotypes of particular groups (e.g., male students' average rating on mathematics is often higher than that on humanities, and vice versa for females) may yield a limited scope of suggested items to a certain group of users. Our main contribution lies in the introduction of a novel fairness notion (that we call \emph{equal experience}), which can serve to regulate such unfairness in the presence of biased data. The notion captures the degree of the equal experience of item recommendations across distinct groups. We propose an optimization framework that incorporates the fairness notion as a regularization term, as well as introduce computationally-efficient algorithms that solve the optimization. Experiments on synthetic and benchmark real datasets demonstrate that the proposed framework can indeed mitigate such unfairness while exhibiting a minor degradation of recommendation accuracy.
\end{abstract}

\section{Introduction}
\label{sec:introduction}
Recommender systems are everywhere, playing a crucial role to support decision making and to decide what we experience in our daily life. One recent challenge concerning \emph{fairness} arises when the systems are built upon biased historical data. Biased data due to polarized preferences of particular groups for certain items may often yield limited recommendation service. For instance, if female students exhibit high ratings on literature subjects and less interest in math and science relative to males, the subject recommender system trained based on such data may provide a narrow scope of recommended subjects to the female group, thereby yielding \emph{unequal} experience. This unequal experience across groups may result in amplifying the gender gap issue in science, technology, engineering, and mathematics (STEM) fields. 

Among various works for fair recommender systems~\cite{yao2017, li2021, kamishima2017, xiao2017, beutel2019, burke2017}, one recent and most relevant work is~\cite{yao2017}. They focus on a scenario in which unfairness occurs mainly due to distinct recommendation accuracies across different groups. They propose novel fairness measures that quantify the degree of such unfairness via the difference between recommendation accuracies, and also develop an optimization framework that well trades the fairness measures against the average  accuracy. However, it comes with a challenge in ensuring fairness w.r.t. the \emph{unequal experience}. This is because similar accuracy performances between different groups do not guarantee a variety of recommendations to an underrepresented group with historical data bearing low preferences and/or scarce ratings for certain items. For instance, in the subject recommendation, the fairness notion may not serve properly, as long as female students exhibit low ratings (and/or lack of ratings) on math and science subjects due to societal/cultural influences (and/or sampling biases). Furthermore, if the recommended items are selected only according to the overall preference, the biased preference for a specific item group will further increase, and the exposure to the unpreferred item group will gradually decrease.

\textbf{Contribution:} In an effort to address the challenge, we introduce a new fairness notion that we call \emph{equal experience}. At a high level, the notion represents how equally various items are suggested even for an underrepresented group preserving such biased historical data. Inspired by an information-theoretic notion \emph{``mutual information''}~\cite{cover1999} and its key property \emph{``chain rule''}, we quantify our notion so as to control the level of \emph{independence} between preference predictions and items for \emph{any group} of users. Specifically, the notion encourages prediction $\widetilde{Y}$ (e.g., 1 if a user prefers an item; 0 otherwise) to be independent of the following two: (i) user group $Z_{\sf user}$ (e.g., 0 for male; and 1 for female); and (ii) item group $Z_{\sf item}$ (e.g., 0 for mathematics; and 1 for literature). In other words, it promotes $\widetilde{Y}\perp (Z_{\sf user}, Z_{\sf item})$; which in turns ensures all of the following four types of independence that one can think of: $\widetilde{Y}\perp Z_{\sf item}$, $\widetilde{Y}\perp Z_{\sf user}$, $\widetilde{Y}\perp Z_{\sf item}|Z_{\sf user}$, and $\widetilde{Y}\perp Z_{\sf user}|Z_{\sf item}$. This is inspired by the fact that mutual information being zero is equivalent to the independence between associated random variables, as well as the chain rule:
\begin{equation}
\label{eq:chain_rule(intro)}
\begin{aligned}
    I(\widetilde{Y};Z_{\sf user}, Z_{\sf item}) 
    &= I(\widetilde{Y};Z_{\sf item}) + I(\widetilde{Y};Z_{\sf user}|Z_{\sf item})\\
    &= I(\widetilde{Y};Z_{\sf user}) + I(\widetilde{Y};Z_{\sf item}|Z_{\sf user}).
\end{aligned}    
\end{equation}
% \begin{align}
% \label{eq:chain_rule(intro)}
% \begin{split}
%     I(\widetilde{Y};Z_{\sf user}, Z_{\sf item}) 
%     &= I(\widetilde{Y};Z_{\sf item}) + I(\widetilde{Y};Z_{\sf user}|Z_{\sf item})\\
%     &= I(\widetilde{Y};Z_{\sf user}) + I(\widetilde{Y};Z_{\sf item}|Z_{\sf user}).
% \end{split}    
% \end{align}

See Section~\ref{sec:new_fairness} for details. The higher independence, the more diverse recommendation services are offered for every group. We also develop an optimization framework that incorporates the quantified notion as a regularization term into a conventional optimization in recommender systems (e.g., the one based on matrix completion~\cite{koren2008, koren2009}). Here one noticeable feature of our framework is that the fairness performances w.r.t. the above \emph{four} types of independence conditions can be gracefully controlled via a \emph{single unified} regularization term. This is in stark contrast to prior works~\cite{yao2017, li2021, kamishima2017, mehrotra2018}, each of which promotes only one independence condition or two via two separate regularization terms. See below {\bf Related works} for details. In order to enable an efficient implementation of the fairness constraint, we employ recent methodologies developed in the context of fair classifiers, such as the ones building upon kernel density estimation~\cite{cho2020kde}, mutual information~\cite{zhang2018mitigating, kamishima2012, cho2020mi}, or covariance~\cite{zafar2017a, zafar2017b}. We also conduct extensive experiments both on synthetic and two benchmark real datasets: MovieLens 1M~\cite{data_movielens} and Last FM 360K~\cite{data_lastfm}. As a result, we first identify two primary sources of biases that incur \emph{unequal experience}: population imbalance and observation bias~\cite{yao2017}. In addition, we demonstrate that our fairness notion can help improve the fairness measure w.r.t. \emph{equal experience} (to be defined in Section~\ref{sec:new_fairness}; see Definition~\ref{def:EE}) while exhibiting a small degradation of recommendation accuracy. Furthermore, we provide an extension of our fairness notion to the context of top-$K$ recommendation from an end-ranked list. We also demonstrate the effectiveness of the proposed framework in top-$K$ recommendation setting.

\textbf{Related works:} In addition to~\cite{yao2017}, numerous fairness notions and algorithms have been proposed for fair recommender systems~\cite{xiao2017, beutel2019, singh2018, zehlike2017, narasimhan2020, biega2018,li2021,kamishima2017, mehrotra2018, schnabel2016}.~\cite{xiao2017} develop fairness notions that encourage similar recommendations for users within the same group.~\cite{beutel2019} consider similar metrics as that in~\cite{yao2017} yet in the context of pairwise recommender systems wherein pairewise preferences are given as training data.~\cite{li2021} propose a fairness measure that quantifies the irrelevancy of preference predictions to user groups, like \emph{demographic parity} in the fairness literature~\cite{feldman2015, zafar2017a, zafar2017b}. Specifically, they consider the independence condition between prediction $\widetilde{Y}$ and user group $Z_{\sf user}$: $\widetilde{Y}\perp Z_{\sf user}$. Actually this was also considered as another fairness measure in~\cite{yao2017}. Similarly, other works with a different direction consider the similar notion concerning the independence w.r.t. item group $Z_{\sf item}$: $\widetilde{Y}\perp Z_{\sf item}$~\cite{kamishima2017,  singh2018, biega2018}.~\cite{mehrotra2018} incorporate both measures to formulate a multi-objective optimization. In Section~\ref{sec:fair_measure}, we will elaborate on why the above prior fairness notions cannot fully address the challenge w.r.t. \emph{unequal} experience. 

There has been a proliferation of fairness notions in the context of fair classifiers: (i) group fairness~\cite{feldman2015, zafar2017b, hardt2016, woodworth2017}; (ii) individual fairness~\cite{dwork2012, garg2018}; (iii) causality-based fairness~\cite{kusner2017,  nabi2018,  russell2017, wu2019, zhang2018eo, zhang2018decision}. Among various prominent group fairness notions, \emph{demographic parity} and \emph{equalized odds} give an inspiration to our work in the process of applying the chain rule, reflected in~\eqref{eq:chain_rule(intro)}. Concurrently, a multitude of fairness algorithms have been developed with the use of covariance~\cite{zafar2017a, zafar2017b}, mutual information~\cite{zhang2018mitigating, kamishima2012, cho2020mi}, kernel density estimation~\cite{cho2020kde} or R\'{e}nyi correlation~\cite{mary2019} to name a few. In this work, we also demonstrate that our proposed framework (to be presented in Section~\ref{sec:proposed_framework}) embraces many of these approaches; See Remark~\ref{remark:other_choices} for details.

\section{Problem Formulation}
\label{sec:problem_formulation}
As a key technique for operating recommender systems, we consider collaborative filtering which estimates user ratings on items. We first formulate an optimization problem building upon one prominent approach, \emph{matrix completion}. We then introduce a couple of fairness measures proposed by recent prior works~\cite{yao2017, li2021, kamishima2017}, and present an extended optimization framework that incorporates the fairness measures as regularization terms.
 
\subsection{Optimization based on matrix completion}
\label{sec:opt_matrix_completion}
As a well-known approach for operating recommender systems, we consider matrix completion~\cite{fazel2002, koren2009, candes2009}. Let $M \in \mathbb{R}^{n \times m}$ be the ground-truth rating matrix where $n$ and $m$ denote the number of users and items respectively. Each entry, denoted by $M_{ij}$, can be of any type. It could be binary, five-star rating, or any real number. Denote by $\Omega$ the set of observed entries of $M$. For simplicity, we assume noiseless observation. 
%that a partially observed matrix, say $Y \in \mathbb{R}^{n \times m}$, is noiseless, i.e., $Y_{ij} = M_{ij}$ if $(i,j) \in \Omega$. Otherwise, we set $Y_{ij} = *$ where $*$ indicates no observation. 
Denote by $\widehat{M} \in \mathbb{R}^{n \times m}$ an estimate of the rating matrix. 
%Let $\widehat{\mathbf{Y}}$ be the partially-observed version w.r.t. $\widehat{\mathbf{X}}$: $\widehat{Y}_{ij}=\widehat{X}_{ij}$ if $X_{ij}$ is observed; $\widehat{Y}_{ij}=0$ otherwise.

Matrix completion can be done via the rank minimization that exploits the low-rank structure of the rating matrix. However, since the problem is NP-hard~\cite{fazel2002}, we consider a well-known relaxation approach that intends to minimize instead the squared error between $M$ and $\widehat{M}$ in the observed entries:
\begin{equation}
\label{eq:MC_base}
    \min_{\widehat{M}} \sum_{(i,j) \in \Omega} ( M_{ij} - \widehat{M}_{ij} )^2.
\end{equation}
%where $ \| \cdot \|_F$ denotes the Frobenius norm. 
There are two well-known approaches for solving the optimization in~\eqref{eq:MC_base}: (i) matrix factorization~\cite{abadir2005, koren2009}; and (ii) neural-net-based parameterization~\cite{salakhutdinov2007, sedhain2015, he2017}. Matrix factorization assumes a certain structure on the rating matrix: $M = LR $ where $L \in \mathbb{R}^{n \times r}$ and $R \in \mathbb{R}^{r \times m}$. One natural way to search for optimal $L^*$ and $R^*$ is to apply gradient descent~\cite{robbins1951} w.r.t. all of the $L_{ij}$'s and $R_{ij}$'s, although it does not ensure the convergence of the optimal point due to non-convexity. The second approach is to parameterize $\widehat{M}$ via neural networks such as restricted Boltzmann machine~\cite{salakhutdinov2007} and autoencoder~\cite{sedhain2015, lee2018}. For instance, one may employ an autoencoder-type neural network which outputs a completed matrix $\widehat{M}$ fed by the partially-observed version of $M$. For a user-based autoencoder~\cite{sedhain2015}, an observed \emph{row} vector of $M$ is fed into the autoencoder, while an observed \emph{column} vector serves as an input for an item-based autoencoder~\cite{sedhain2015}. In this work, we consider the two approaches in our experiments: matrix factorization with gradient descent; and autoencoder-based parameterization.

One common way to promote a \emph{fair} recommender system is to incorporate a fairness measure, say ${\cal L}_{\sf fair}$ (which we will relate to an estimated matrix $\widehat{M}$), as a regularization term into the above base optimization in~\eqref{eq:MC_base}: 
\begin{equation}
\label{eq:MC_fair}
     \min_{\widehat{M}} \; (1-\lambda) \sum_{(i,j) \in \Omega} ( M_{ij} - \widehat{M}_{ij} )^2 + \lambda\cdot \mathcal{L}_{\sf fair} 
\end{equation}
where $\lambda \in [0,1]$ denotes a normalized regularization factor that balances prediction accuracy against the fairness constraint. For the fairness-regularization term ${\cal L}_{\sf fair}$, several fairness measures have been introduced.

\subsection{Fairness measures in prior works~\cite{yao2017, kamishima2017, li2021}}
\label{sec:fair_measure}

We list three of them, which are mostly relevant to our framework to be presented in Section~\ref{sec:proposed_framework}. For illustrative purpose, we will explain them in a simple setting where there are two groups of users, say the male group ${\cal M}$ and the female group ${\cal F}$. The first is \emph{value unfairness} proposed by~\cite{yao2017}. It quantifies the difference between prediction errors across the two groups of users over the entire items:
\begin{align} 
\label{eq:VAL}
\begin{split}
    {\sf VAL} := \frac{1}{m}\sum_{j=1}^m \bigg \vert \underbrace{ \frac{1}{|\mathcal{M}_{\Omega}|} \sum\limits_{(i,j) \in \Omega: i \in \mathcal{M} } ( M_{ij} - \widehat{M}_{ij} )}_{\textrm{prediction error w.r.t. } {\cal M}}\\
    - \underbrace{ \frac{1}{|\mathcal{F}_{\Omega}|} \sum \limits_{(i,j) \in \Omega: i \in \mathcal{F}} ( M_{ij} - \widehat{M}_{ij} ) }_{ \textrm{prediction error w.r.t. } {\cal F}  }  \bigg \vert
\end{split}
\end{align}
where ${\cal M}_{\Omega}$ and ${\cal F}_{\Omega}$ denote the male and female group w.r.t. observed entries, respectively. While the measure promotes fairness w.r.t. \emph{prediction accuracy} across distinct groups, it may not ensure fairness w.r.t. the diversity of recommended items to users. To see this clearly, consider an extreme scenario in which the ground truth rating is very small $M_{i j^*} \approx 0$ for a certain item $j^*$ (say science subject) for all $i \in {\cal F}$. In this case, minimizing {\sf VAL} may encourage $\widehat{M}_{i j^*} \approx 0$ for all $i \in {\cal F}$. This then incurs almost no recommendation of the science subject to the females, thus giving no opportunity to experience the subject. This motivates us to propose a new fairness measure (to be presented in Section~\ref{sec:new_fairness}) that helps mitigate such unfairness.

\begin{figure}[t]
\begin{center}
    \includegraphics[height=7cm]{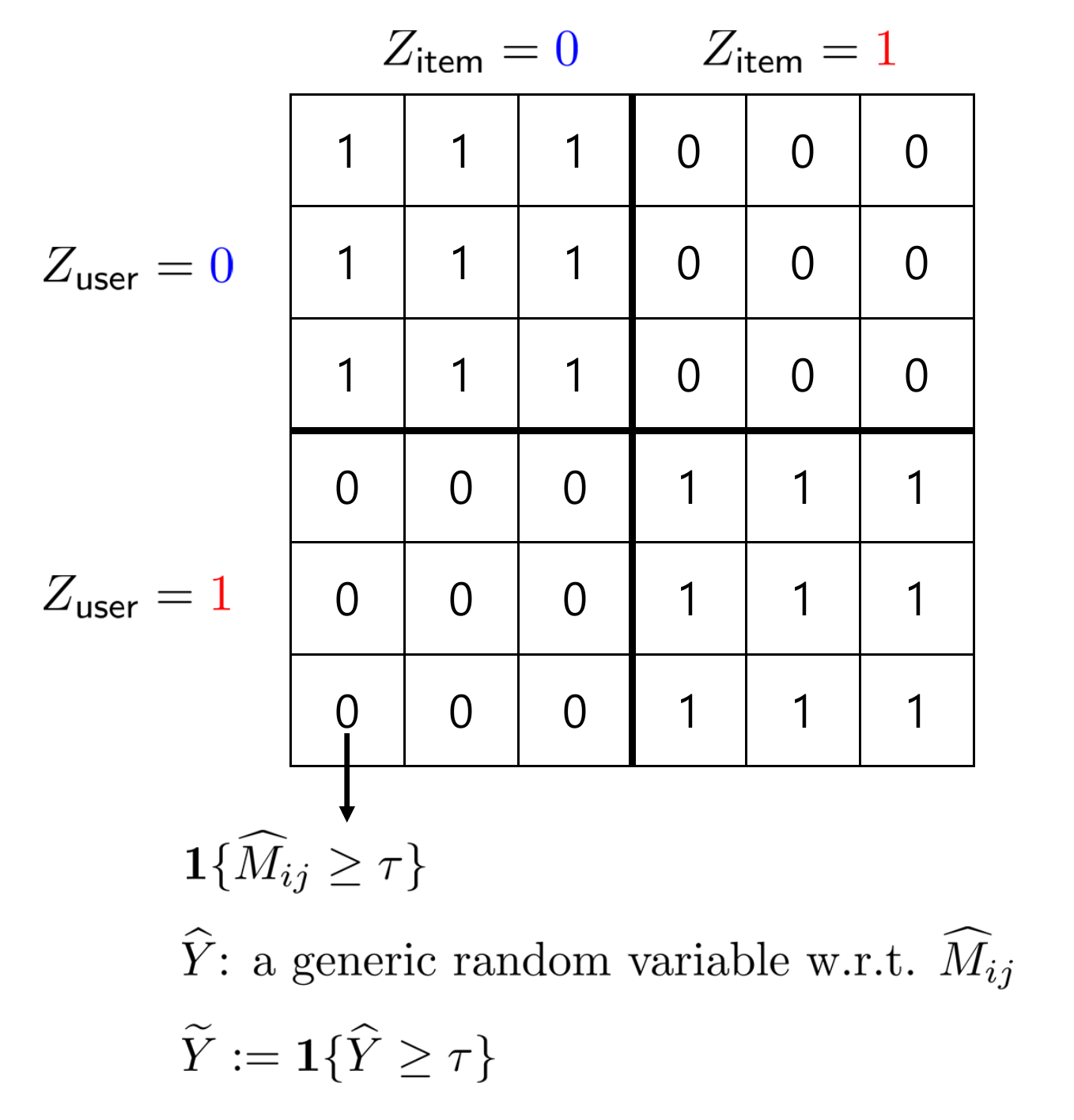}
    \caption{An example in which $\widetilde{Y} \perp Z_{\sf item}$ but $\widetilde{Y} \not\perp Z_{\sf item} | Z_{\sf user}$. Here $\widetilde{Y} := {\bf 1} \{ \widehat{Y} \geq \tau \}$; $\widehat{Y}$ is a generic random variable w.r.t. estimated ratings $\widehat{M}_{ij}$'s; and $\tau$ indicates a certain threshold. The $(i,j)$ entry of an estimated preference matrix with 6 users (row) and 6 items (column) indicates ${\bf 1} \{ \widehat{M}_{ij} \geq \tau \}$.}
    \label{fig:1}
\end{center}
\end{figure}

On the other hand,~\cite{kamishima2017} introduce another fairness measure, which bears a similar spirit to demographic parity in the fairness literature~\cite{feldman2015, zafar2017a, zafar2017b}. The measure, named \emph{Calders and Verwer's discrimination score} (${\sf CVS}$), quantifies the level of irrelevancy between preference predictions and item groups. To describe it in detail, let us introduce some notations. Let $Z_{\sf item}$ be a sensitive attribute w.r.t. item groups, e.g., $Z_{\sf item} = 0$ (literature) and $Z_{\sf item} = 1$ (science). Let $\widehat{Y}$ be a generic random variable w.r.t. estimated ratings $\widehat{M}_{ij}$'s. To capture the preference prediction, let us consider a simple binary preference setting in which $\widetilde{Y} := {\bf 1} \{ \widehat{Y} \geq \tau \}$ where $\tau$ indicates a certain threshold. Specializing the measure into the one like demographic parity, it can be quantified as:
\begin{equation}
\label{eq:CVS}
\resizebox{.9\hsize}{!}{$
    {\sf CVS} := | \mathbb{P} (\widetilde{Y} = 1 | Z_{\sf item} = 1) - \mathbb{P} (\widetilde{Y} = 1 | Z_{\sf item} = 0) |.$
}
\end{equation}
Minimizing the measure encourages the independence between $\widetilde{Y}$ and $Z_{\sf item}$, thereby promoting the same rating statistics across different groups. However, it does not necessarily ensure the same statistics when we focus on a \emph{certain} group of users. It guarantees the independence only in the \emph{average} sense. To see this clearly, consider a simple scenario in which there are two groups of users, say female and male. Let $Z_{\sf user}$ be another sensitive attribute w.r.t. user groups, e.g., $Z_{\sf user} = 0$ (female) and $Z_{\sf user} = 1$ (male). Fig.~\ref{fig:1} illustrates a concrete example where $\widetilde{Y}$ is independent of $Z_{\sf item}$. 

% \begin{figure}[t]
% \begin{center}
%     \subfloat{\includegraphics[height=5cm]{fig1_R3.pdf}}
%   % \hspace{0.5cm}
%   % \subfloat{\includegraphics[height=4.3cm]{fig1_2.pdf}}
%     \caption{An example in which $\widetilde{Y} \perp Z_{\sf item}$ but $\widetilde{Y} \not\perp Z_{\sf item} | Z_{\sf user}$. Here $\widetilde{Y} := {\bf 1} \{ \widehat{Y} \geq \tau \}$; $\widehat{Y}$ is a generic random variable w.r.t. estimated ratings $\widehat{M}_{ij}$'s; and $\tau$ indicates a certain threshold. The $(i,j)$ entry of an estimated preference matrix with 6 users (row) and 6 items (column) indicates ${\bf 1} \{ \widehat{M}_{ij} \geq \tau \}$.} 
%     %(Right) An example where $\widehat{Y} \perp Z_{\sf user} | Z_{\sf item}$ but $\widehat{Y} \not\perp Z_{\sf user}$.}
%     \label{fig:1}
% \end{center}
% \end{figure}
Notice that the number of 1's w.r.t. $Z_{\sf item}=0$ (over the entire user groups) is the same as that w.r.t. $Z_{\sf item}=1$. However, focusing on a certain user group, say $Z_{\sf user} =0$, $\widetilde{Y}$ is highly correlated with $Z_{\sf item}$. Observe in the case  $Z_{\sf user}= 0$ that the number of 1's is 9 for $Z_{\sf item}= 0$, while it reads 0 for $Z_{\sf item} = 1$.

\cite{li2021} consider a similar measure, named \emph{user-oriented group fairness} (${\sf UGF}$), yet which targets the independence w.r.t. \emph{user} groups. Similar to ${\sf CVS}$, we can define it by replacing $Z_{\sf item}$ with $Z_{\sf user}$ in~\eqref{eq:CVS}:
\begin{equation}
\label{eq:UGF}
\resizebox{.9\hsize}{!}{$
    {\sf UGF} := | \mathbb{P} (\widetilde{Y} = 1 | Z_{\sf user} = 1) - \mathbb{P} (\widetilde{Y} = 1 | Z_{\sf user} = 0) |.$
}
\end{equation}
However, by symmetry, the high correlation issue discussed via Fig.~\ref{fig:1} still arises.

\section{Proposed Framework}
\label{sec:proposed_framework}

We first propose new fairness measures that can regulate fairness w.r.t. the opportunity to experience inherently-low preference items, as well as address the high correlation issue discussed as above. We then develop an integrated optimization framework that unifies the fairness measures as a single regularization term. Finally we introduce concrete methodologies that can implement the proposed optimization. 

\subsection{New fairness measures}
\label{sec:new_fairness}

The common limitation of the prior fairness measures~\cite{yao2017, kamishima2017, li2021} is that the independence between preference predictions and item groups may not be guaranteed for a \emph{certain} group of users. This motivates us to consider the \emph{conditional} independence as a new fairness notion, formally defined as below. 
 
\begin{definition}[Equalized Recommendation]
\label{def:ER}
A recommender system is said to respect ``equalized recommendation'' if its prediction $\widetilde{Y}$ is independent of item's sensitive attribute $Z_{\sf item}$ given user's sensitive attribute $Z_{\sf user}$: $\widetilde{Y}\perp Z_{\sf item}|Z_{\sf user}$.
\end{definition}
Inspired by the quantification methods w.r.t. \emph{equalized odds} in the fairness literature~\cite{jiang2019, donini2018, hardt2016, woodworth2017}, we quantify the new notion via:
\begin{align}\label{DER}
    \begin{split}
    {\sf DER}:=\sum_{z_1 \in\mathcal{Z}_{\sf user}}\sum_{z_2 \in\mathcal{Z}_{\sf item}}
    \Big \vert &\mathbb{P} (\widetilde{Y}=1|Z_{\sf user}=z_1) \\ 
    -&\mathbb{P}(\widetilde{Y}=1|Z_{\sf item} = z_2,Z_{\sf user}=z_1 ) \Big \vert,
    \end{split}
\end{align}
for arbitrary alphabet sizes $|\mathcal{Z}_{\sf user}|$ and $|\mathcal{Z}_{\sf item}|$. Here ${\sf DER}$ stands for the difference w.r.t. two interested probabilities that arise in equalized recommendation, and this naming is similar to those in prior fairness metrics~\cite{donini2018, jiang2019}. It captures the degree of violating equalized recommendation via the difference between the conditional probability and its marginal given $Z_{\sf user}$. Notice that the minimum ${\sf DER} = 0$ is achieved under ``equalized recommendation''. One may consider another measure which takes ``max'' operation instead of ``$\sum$'' in~\eqref{DER} or a different measure based on the \emph{ratio} of the two associated probabilities. We focus on ${\sf DER}$ in~\eqref{DER} for tractability of an associated optimization problem that we will explain in Section~\ref{sec: proposed_opt}.

The constraint of ``equalized recommendation'' encourages the same prediction statistics of items for \emph{every} user group, thereby promoting the equal chances of experiencing a variety of items for all individuals. However, the notion comes with a limitation. The limitation comes from the fact that conditional independence does not necessarily imply independence~\cite{cover1999}:
\begin{align}
\label{eq:ER_limitation}
\widetilde{Y} \perp Z_{\sf item} | Z_{\sf user} \;\; \centernot\Longrightarrow \;\; \widetilde{Y} \perp Z_{\sf item}.
\end{align}
Actually, the ultimate goal of a fair recommender system is to ensure all of the following four types of independence:
\begin{align}
\begin{split}
 \widetilde{Y} \perp Z_{\sf item},\quad \widetilde{Y} \perp Z_{\sf user} | Z_{\sf item}, \\ \widetilde{Y} \perp Z_{\sf user}, \quad \widetilde{Y} \perp Z_{\sf item} | Z_{\sf user}.
\end{split}
\end{align}
One natural question that arises is then: What is a proper fairness notion which allows us to respect all of the above four conditions preferably in one shot? 
 
In an attempt to succinctly represent all of the four conditions, we invoke an information-theoretic notion, \emph{mutual information}~\cite{cover1999}. One key property of mutual information, called the chain rule, gives an insight:
\begin{align}
\begin{split}
    I(\widetilde{Y};Z_{\sf user}, Z_{\sf item}) 
    &= I(\widetilde{Y};Z_{\sf item}) + I(\widetilde{Y};Z_{\sf user}|Z_{\sf item})\\
    &= I(\widetilde{Y};Z_{\sf user}) + I(\widetilde{Y};Z_{\sf item}|Z_{\sf user}).
\end{split}
\end{align}
From this, we can readily see that
\begin{align}
\begin{split}
\label{eq:key_observation}
  &I(\widetilde{Y};Z_{\sf user}, Z_{\sf item})=0 \;\;\\ &\Longrightarrow \;\;  I(\widetilde{Y};Z_{\sf item})=0, \; I(\widetilde{Y};Z_{\sf user}|Z_{\sf item})=0,\\
    &\quad\qquad I(\widetilde{Y};Z_{\sf user})=0, \; I(\widetilde{Y};Z_{\sf item}|Z_{\sf user})=0.
\end{split}
\end{align}
This is due to the non-negativity property of mutual information. The key observation in~\eqref{eq:key_observation} motivates us to propose a new fairness notion that we call \emph{equal experience}. 
\begin{definition}[Equal Experience]
\label{def:EE}
A recommender system is said to respect ``equal experience'' if its prediction $\widetilde{Y}$ is independent of both $Z_{\sf item}$ and $Z_{\sf user}$: $\widetilde{Y}\perp (Z_{\sf item},Z_{\sf user})$.
\end{definition}
Similar to ${\sf DER}$, we also quantify the notion as the difference between the conditional probability and its marginal:
\begin{align}\label{def:DEE}
\begin{split}
    \textsf{DEE}:=\sum_{z_1 \in \mathcal{Z}_{\sf user}}\sum_{z_2 \in \mathcal{Z}_{\sf item}}
    \Big\vert &\mathbb{P} (\widetilde{Y}=1)\\
    -&\mathbb{P} (\widetilde{Y}=1|Z_{\sf item}=z_2, Z_{\sf user}=z_1 )\Big\vert,
\end{split}
\end{align}
for arbitrary alphabet sizes $|\mathcal{Z}_{\sf user}|$ and $|\mathcal{Z}_{\sf item}|$. We also coin the similar naming: ${\sf DEE}$ (difference w.r.t. two interested probabilities that arise in equal experience).

\subsection{Fairness-regularized optimization}
\label{sec: proposed_opt}

Taking ${\sf DEE}$ as the fairness-regularization term ${\cal L}_{\sf fair}$ in the focused framework (\eqref{eq:MC_fair}), we get:
\begin{equation}
\label{eq:MC_DEE}
     \min_{\widehat{M}} \; (1- \lambda) \sum_{(i,j) \in \Omega} ( M_{ij} - \widehat{M}_{ij} )^2 + \lambda\cdot {\sf DEE} 
\end{equation}
where $\lambda \in [0,1]$ denotes a normalized regularization factor. Here one challenge that arises in~\eqref{eq:MC_DEE} is that expressing ${\sf DEE}$ in terms of an optimization variable $\widehat{M}$ is not that straightforward.

To overcome the challenge, we take the kernel density estimation (KDE) technique~\cite{cho2020kde} which enables faithful quantification of fairness-regularization terms. One key benefit of the KDE approach is that the computed measures based on KDE is differentiable w.r.t. model parameters, thus enjoying a family of gradient-based optimizers~\cite{geron2019, kingma2014}. Since the problem context where the KDE technique~\cite{cho2020kde} was introduced is different from ours, we describe below details on the technique, tailoring it to our framework. 

{\bf Implementation of ${\sf DEE}$ via the KDE technique~\cite{cho2020kde}:} We first parameterize prediction output $\widehat{M}$ via matrix factorization or a neural network. Let $w$ be a collection of parameters w.r.t. $\widehat{M}$. It could be a collection of matrix entries of $L$ and $R$ when employing matrix factorization $\widehat{M} = LR$. Or it could be a collection of neural network parameters in the latter case.   

The key idea of the KDE technique is to \emph{approximate} the interested probability distributions via kernel density estimator defined below:
\begin{definition}[Kernel Density Estimator (KDE)~\cite{davis2011}]
Let $(\hat{y}^{(1)},\ldots, \hat{y}^{(s)})$ be i.i.d. examples drawn from a distribution with an unknown density $f$. Its KDE is defined as: $\widehat{f}(\hat{y}) := \frac{1}{s h}\sum_{i=1}^s f_k\left(\frac{\hat{y}-\hat{y}^{(i)}}{h}\right)$
where $f_k$ is a kernel function (e.g., Gaussian kernel function~\cite{davis2011}) and $h>0$ is a smoothing parameter called bandwidth. 
\end{definition}
For ${\sf DEE}$, the interested probability distributions are $\mathbb{P}(\widetilde{Y} =1)$ and  $\mathbb{P}(\widetilde{Y} =1 | Z_{\sf item} = z_2, Z_{\sf user} = z_1)$. Let us first consider $\mathbb{P}(\widetilde{Y} =1)$. Remember $\widetilde{Y} := \mathbf{1} \{ \widehat{Y} \geq \tau \}$, so $\widehat{Y}$ should be taken into consideration initially. Using the KDE, we can estimate the probability density function of $\widehat{Y}$, say $f_{\widehat{Y}} (\hat{y})$:
\begin{equation}
\label{eq:kernel}
    \widehat{f}_{\widehat{Y}}(\hat{y}) = \frac{1}{n m h}\sum_{i =1 }^n \sum_{j=1}^m f_k\left({\frac{\hat{y}-\widehat{M}_{ij} }{h}}\right).
\end{equation}
This together with $\widetilde{Y} := \mathbf{1} \{ \widehat{Y} \geq \tau \}$ gives:
\begin{align}
  \widehat{ \mathbb{P} } (\widetilde{Y} = 1) 
  %&= \hat{\Pr}\left(\widehat{Y}\geq\tau|Z=z\right)
  =\int_{\tau}^{\infty}\widehat{f}_{\widehat{Y}}(\hat{y})d\hat{y}=\frac{1}{n m }\sum_{i = 1}^n \sum_{j=1}^m 
  F_k\left({\frac{\tau-\widehat{M}_{ij} }{h}}\right)
\end{align}
where $F_k(\hat{y}):=\int_{\hat{y}}^{\infty}f_k(t)dt$. Since the approach relies upon a family of gradient-based optimizers, the gradients of $\widehat{ \mathbb{P} } (\widetilde{Y} = 1)$ and $\widehat{\mathbb{P}}(\widetilde{Y} =1 | Z_{\sf item} = z_2, Z_{\sf user} = z_1)$ need to be computed explicitly. Using the technique in~\cite{cho2020kde} (Proposition 1 therein), we can readily approximate $\nabla_w\textsf{DEE}$. See Appendix~\ref{sec:appendix_kde} for details.

% \begin{remark}[Other choices for a measure of ``equal experience''] \label{remark:other_choices} Instead of ${\sf DEE}$, one can resort to other measures based on prominent tools employed in the fairness literature: covariance~\cite{zafar2017a, zafar2017b}, mutual information~\cite{zhang2018mitigating, kamishima2012, cho2020mi}, Wasserstein distance~\cite{jiang2020} and R\'{e}nyi correlation~\cite{mary2019}. We leave a more detailed explanation in Appendix~\ref{sec:appendix_measure}.
% $\blacksquare $
% \end{remark}

\begin{remark}[Other choices for a measure of ``equal experience''] \label{remark:other_choices} Instead of ${\sf DEE}$, one can resort to other measures based on prominent tools employed in the fairness literature: covariance~\cite{zafar2017a, zafar2017b}, mutual information~\cite{zhang2018mitigating, kamishima2012, cho2020mi}, Wasserstein distance~\cite{jiang2020} and R\'{e}nyi correlation~\cite{mary2019}. For instance, the covariance-based approach allows us to take 
 ${\cal L}_{\sf fair}$ as:
\begin{align}
{\cal L}_{\sf fair} = \mathbb{E} \left[  (\widehat{Y} - \mathbb{E} [\widehat{Y}]) (Z - \mathbb{E}[Z]) \right],
\end{align}
% \qquad \textrm{where } Z:= (Z_{\sf item}, Z_{\sf user} ).
where $Z:= (Z_{\sf item}, Z_{\sf user} )$. Here we use $\widehat{Y}$ instead of $\widetilde{Y}$, as $\widetilde{Y}$ incurs non-differentiability, hindering implementation. In the case of mutual information, one can take ${\cal L}_{\sf fair}$ as:
\begin{align}
{\cal L}_{\sf fair} = I (\widehat{Y}; Z_{\sf item}, Z_{\sf user} ) \geq I (\widetilde{Y}; Z_{\sf item}, Z_{\sf user}). 
\end{align}
Again, for ease of implementation, we employ $\widehat{Y}$. This choice is also relevant because it serves as an upper bound of $I (\widetilde{Y}; Z_{\sf item}, Z_{\sf user})$. Notice that $\widetilde{Y}$ is a function of $\widehat{Y}$. Reducing $I (\widehat{Y}; Z_{\sf item}, Z_{\sf user} )$ yields the minimization of the interested quantity $I (\widetilde{Y}; Z_{\sf item}, Z_{\sf user})$. For faithful implementation of $I (\widehat{Y}; Z_{\sf item}, Z_{\sf user} )$, we may employ the variational optimization technique in~\cite{zhang2018mitigating, cho2020mi} to translate it into a function optimization which can also be parameterized. Other choices can also be dealt with properly relying upon the associated techniques in~\cite{jiang2020, mary2019}. 
 $\blacksquare $
\end{remark}

\begin{figure*}[t]
    \centering  
    \subfloat{\includegraphics[height=4.3cm]{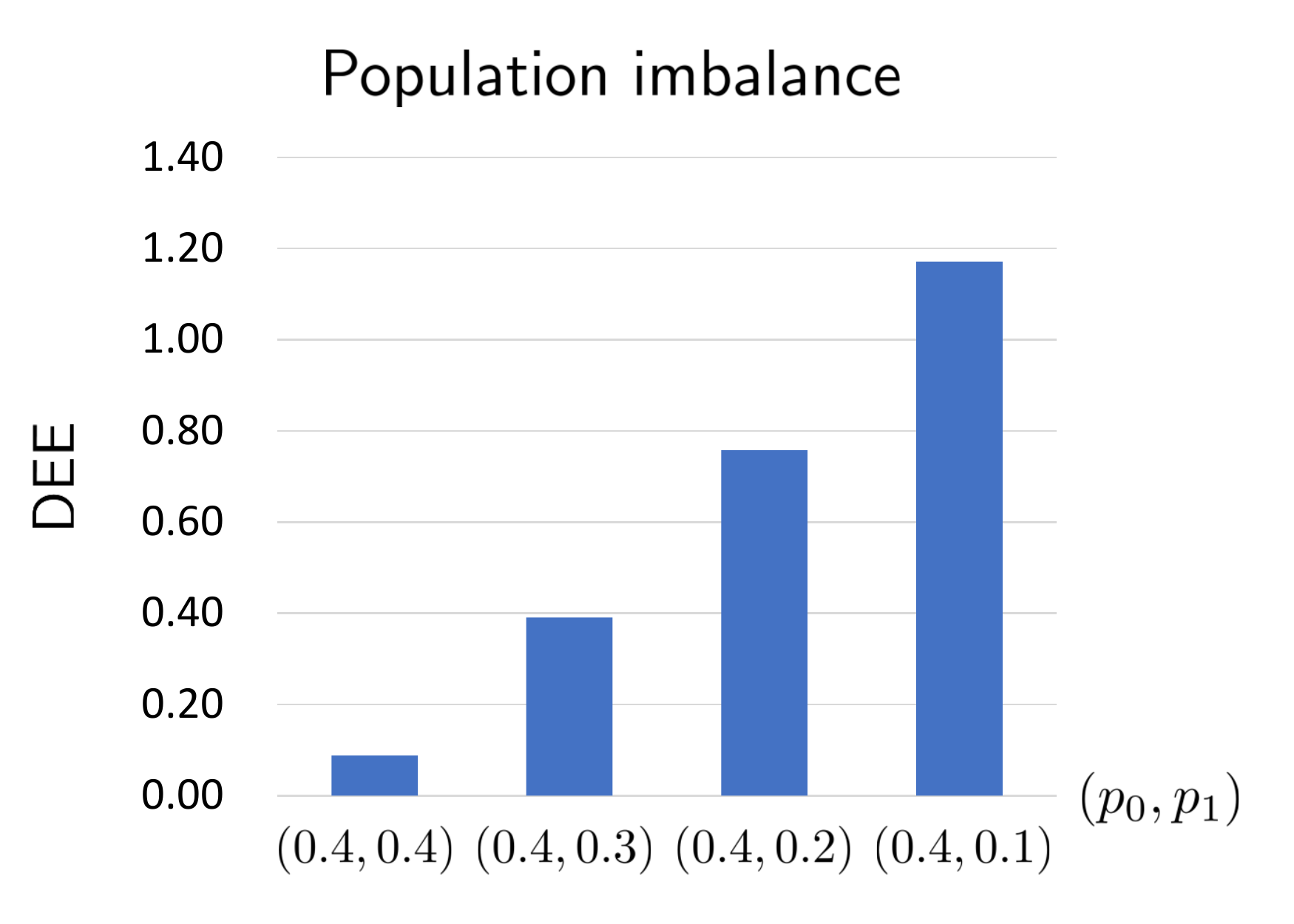}}
    \hspace{0.5cm}
    \subfloat{\includegraphics[height=4.3cm]{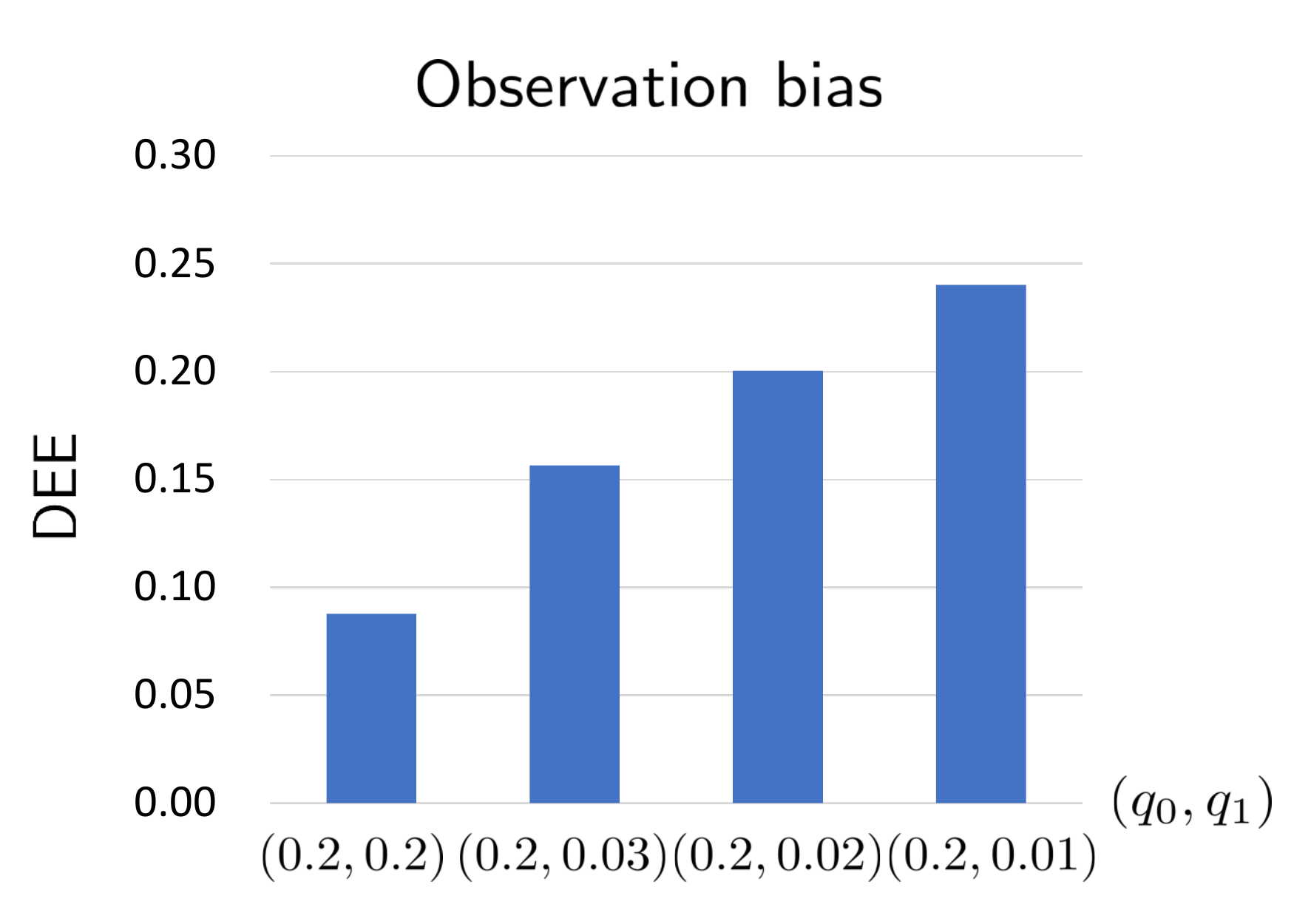}}
    \caption{ (Left) $\sf DEE$ as a function of $(p_0, p_1)$ which controls the degree of population imbalance while fixing $q_0 = q_1 = 0.2$; (Right) $\sf DEE$ as a function of $(q_0, q_1)$ w.r.t. observation bias while fixing $p_0 = p_1 = 0.4$. Here $p_0 = p_{\mathcal{M}\mathcal{M}'}=p_{\mathcal{F}\mathcal{F}'} $, $p_1 =  p_{\mathcal{M}\mathcal{F}'}=p_{\mathcal{F}\mathcal{M}'}$, $q_0 = q_{\mathcal{M}\mathcal{M}'}=q_{\mathcal{F}\mathcal{F}'} $ and $q_1 =  q_{\mathcal{M}\mathcal{F}'}=q_{\mathcal{F}\mathcal{M}'}$.}
    \label{fig:2}
\end{figure*}

\section{Experiments}
\label{sec: experiments}

We conduct experiments both on synthetic and two benchmark real datasets: MovieLens 1M~\cite{data_movielens} and Last FM 360K~\cite{data_lastfm}. We generate synthetic data so as to pose fairness issues.  Algorithms are implemented in PyTorch~\cite{pytorch}, and experiments are performed on a server with Titan RTX GPUs. All the simulation results (to be reported) are the ones averaged over five trials with distinct random seeds. In Appendix~\ref{sec:appendix_running_times}, we also present the running times of our algorithm and baselines on the synthetic and real datasets. 

\subsection{Synthetic dataset}
\label{sec: synthetic_data}
Here we highlight two types of bias: population imbalance and observation bias~\cite{yao2017}. For illustrative purpose, let us explain them in a simple subject-recommendation example where there are two user groups ($Z_{\sf user} = 0$ for male and $Z_{\sf user} = 1$ for female) and two item groups ($Z_{\sf item}=0$ for science and $Z_{\sf item}=1$ for literature). Population imbalance refers to the difference in the ground-truth preferences between two user groups, e.g., for the science subject, male students exhibit higher ratings relative to females. Observation bias is the one that occurs due to the stereotype formed by societal and cultural influences. To understand what it means, let us consider a scenario where a male student equally likes science and literature subjects. But due to the stereotype that male students prefer science to literature in general, there may be very sparse ratings from male students for literature. The system trained based on such data might incorrectly interpret as if male students dislike literature. Such data is said to have \emph{observation bias}.

We generate synthetic data that bear the two biases in the context of binary ratings, i.e., $M_{ij} \in \{+1\ (\text{like}), -1\ (\text{dislike})\}$. We divide $n$ users into male and female groups each of $\frac{n}{2}$, say $\mathcal{M}$ and $\mathcal{F}$. Items are also divided into two groups of $\frac{m}{2}$, say male-preferred group $\mathcal{M}'$ and female-preferred group $\mathcal{F}'$. 
To account for population imbalance, we first generate the ground-truth rating matrix $M \in \mathbb{R}^{n \times m}$ using the following four probabilities: $\{p_{\mathcal{M}\mathcal{M}'}, p_{\mathcal{M}\mathcal{F}'},  p_{\mathcal{F}\mathcal{M}'}, p_{\mathcal{F}\mathcal{F}'}  \}$ where $p_{\mathcal{M}\mathcal{M}'}$ indicates the probability that a male student likes a male-preferred subject (science). More precisely, for $i \in {\cal M}$ and $j \in {\cal M}'$,
\begin{align}
M_{ij} = \left\{
  \begin{array}{ll}
    +1, & \hbox{w.p. $p_{{\cal M} {\cal M}'}$;} \\
    -1, & \hbox{w.p. $1 - p_{{\cal M} {\cal M}'}$}.
  \end{array}
\right.
\end{align}
Similarly the other probabilities are defined. 
To ensure the low-rank structure, say rank $r$, of the rating matrix, we generate $\frac{r}{2}$ basis rating vectors for male group as per the above preference probabilities, and similarly another set of $\frac{r}{2}$ basis rating vectors is generated for female group. 
Every male student picks one of the $\frac{r}{2}$ basis vectors w.r.t. the male group uniformly at random, and similarly for every female student. This then yields $\text{rank}(\mathbf{M})=r$. 

To control observation bias, we introduce another probability set: $\{q_{\mathcal{M}\mathcal{M}'}, q_{\mathcal{M}\mathcal{F}'},  q_{\mathcal{F}\mathcal{M}'}, q_{\mathcal{F}\mathcal{F}'}  \}$ where $q_{\mathcal{M}\mathcal{M}'}$ denotes the probability that a male student's rating is observed for a male-preferred subject. More precisely, for $i \in {\cal M}$ and $j \in {\cal M}'$,
\begin{align}
(i,j)\in  \left\{
  \begin{array}{ll}
    \Omega, & \hbox{w.p. $q_{{\cal M} {\cal M}'}$;} \\
    \Omega^{c}, & \hbox{w.p. $1 - q_{{\cal M} {\cal M}'}$}.
  \end{array}
\right.
\end{align}
Similarly the other probabilities are defined.
For simplicity, throughout all the synthetic data simulations, we assume a symmetric setting in which $p_{\mathcal{M}\mathcal{M}'}=p_{\mathcal{F}\mathcal{F}'}(=p_0)$, $p_{\mathcal{M}\mathcal{F}'}=p_{\mathcal{F}\mathcal{M}'}(=p_1)$, $q_{\mathcal{M}\mathcal{M}'}=q_{\mathcal{F}\mathcal{F}'}(=q_0)$, and $q_{\mathcal{M}\mathcal{F}'}=q_{\mathcal{F}\mathcal{M}'}(=q_1)$.

% \begin{figure*}[h!]
%     \centering  
%     \subfloat{\includegraphics[height=5cm]{fig2_1.pdf}}
%     \hspace{0.5cm}
%     \subfloat{\includegraphics[height=5cm]{fig2_2.pdf}}
%     \caption{ (Up) $\sf DEE$ as a function of $(p_0, p_1)$ which controls the degree of population imbalance while fixing $q_0 = q_1 = 0.2$; (Down) $\sf DEE$ as a function of $(q_0, q_1)$ w.r.t. observation bias while fixing $p_0 = p_1 = 0.4$. Here $p_0 = p_{\mathcal{M}\mathcal{M}'}=p_{\mathcal{F}\mathcal{F}'} $, $p_1 =  p_{\mathcal{M}\mathcal{F}'}=p_{\mathcal{F}\mathcal{M}'}$, $q_0 = q_{\mathcal{M}\mathcal{M}'}=q_{\mathcal{F}\mathcal{F}'} $ and $q_1 =  q_{\mathcal{M}\mathcal{F}'}=q_{\mathcal{F}\mathcal{M}'}$.}
%     \label{fig:2}
% \end{figure*}

We consider a setting in which $(r, n, m) = (20, 600, 400)$. We leave a more detailed explanation of experiments on the synthetic dataset in Appendix~\ref{sec: appendix_synthetic}. First, we check whether the bias actually incurs unfair recommendations. 
For ease of understanding, we consider two scenarios: (i) population imbalance varies without observation bias; and (ii) observation bias varies without population imbalance. Fig.~\ref{fig:2} (Left) presents the 1st scenario, demonstrating that the fairness performance measured in ${\sf DEE}$ decreases with an increase in population imbalance, controlled by $|p_0 - p_1|$.  Fig.~\ref{fig:2} (Right) considers the 2nd scenario. We see the same trend yet now w.r.t. the variation of observation bias. 

Table~\ref{table:synthetic_MF} presents the prediction error (RMSE) and fairness performances on the synthetic dataset having observation bias $(q_0, q_1) = (0.2,  0.01)$ yet without population imbalance $(p_0,p_1)=(0.4, 0.4)$.  We consider four fairness measures: (i) ${\sf DEE}$ in~\eqref{def:DEE}; (ii) ${\sf VAL}$ in~\eqref{eq:VAL}; (iii) ${\sf UGF}$ in~\eqref{eq:UGF}; (iv) ${\sf CVS}$ in~\eqref{eq:CVS}. We also compare our algorithm with four baselines: (i) unfair (no fairness constraint); (ii) ${\sf VAL}$-based algorithm~\cite{yao2017}; (iii) ${\sf UGF}$-based algorithm~\cite{li2021}; (iv) ${\sf CVS}$-based algorithm~\cite{kamishima2017}. 
Each baseline, say ${\sf VAL}$-based algorithm, achieves the best fairness performance only for ${\sf VAL}$, while it does not work well under the other fairness measures. On the other hand, our algorithm offers great fairness performances for all the measures, except for ${\sf VAL}$, which our framework does not target.

\begin{table*}[!ht]
  \caption{Prediction error (RMSE) and fairness performances on the synthetic dataset preserving observation bias $(q_0,q_1)=(0.2, 0.01)$ while exhibiting no population imbalance $(p_0,p_1)=(0.4, 0.4)$. The boldface indicates the best result and the underline denotes the second best.}
  \label{table:synthetic_MF}
  \centering
  
  \resizebox{6in}{!}{
  \begin{tabular}{cccccc}
    \toprule
    % Measure & RMSE & $\sf DEE$ & $\sf VAL$ & $\sf UGF$ & $\sf CVS$ \\
    Measure & RMSE & $\sf DEE$ & $\sf VAL$ & $\sf UGF$ & $\sf CVS$ \\
    \midrule
    Unfair   & 0.8889 $\pm$ 0.0111 & 0.1201 $\pm$ 0.0405& 0.4646 $\pm$ 0.0108& 0.0120 $\pm$ 0.0050& 0.0118 $\pm$ 0.0047 \\
    Ours (\sf{DEE})   & 0.9020 $\pm$ 0.0081& \textbf{0.0025} $\pm$ \textbf{0.0006} & 0.4609 $\pm$ 0.0076& \underline{0.0006 $\pm$ 0.0004}& \underline{0.0002 $\pm$ 0.0001}\\
    Ours (\sf{DER})  & 0.8887 $\pm$ 0.0042     & {0.0401} $\pm$ {0.0056}   & 0.4540 $\pm$ 0.0110    & {0.0201 $\pm$ 0.0028}  & {0.0002 $\pm$ 0.0002} \\
    $\sf VAL$-based   & 0.8837 $\pm$ 0.0045& 0.1099 $\pm$ 0.0045& \textbf{0.0003} $\pm$ \textbf{6.48e-6} & 0.0090 $\pm$ 0.0004& 0.0088 $\pm$ 0.0015\\
    $\sf UGF$-based   & 0.8961 $\pm$ 0.0067& 0.0144 $\pm$ 0.0144& 0.4709 $\pm$ 0.0182& \textbf{0.0004} $\pm$ \textbf{0.0003} & 0.0217 $\pm$ 0.0027 \\
    $\sf CVS$-based & 0.9003 $\pm$ 0.0061& 0.1390 $\pm$ 0.0413 & 0.4722 $\pm$ 0.0055& 0.0206 $\pm$ 0.0022& \textbf{0.0002} $\pm$ \textbf{0.0001} \\
    \bottomrule
  \end{tabular}}
\end{table*}

\subsection{Real datasets}
\label{sec: real_data}

% We consider two benchmark datasets: MovieLens 1M~\cite{data_movielens}, and Last FM 360K~\cite{data_lastfm}. 
We consider two benchmark datasets: MovieLens 1M~\cite{data_movielens}, and Last FM 360K~\cite{data_lastfm}:
\begin{itemize}
    \item \emph{MovieLens 1M}: The associated task is to predict the movie rating on a 5-star scale. This dataset contains 6,040 users, 3,900 movies, and 1,000,209 ratings, i.e., rating matrix is 4.26\% full.\footnote{http://www.movielens.org/} We divide user and item groups based on gender and genre, respectively. Action, crime, filme-noir, war are selected as male-preferred genre, whereas children, fantasy, musical, romance are selected as female-preferred genre. 
% From the real data, we empirically estimate the interested probabilities w.r.t. population imbalance as: $\widehat{p}_{\mathcal{M}\mathcal{M}'}=0.627,\ \widehat{p}_{\mathcal{M}\mathcal{F}'}=0.517,\  \widehat{p}_{\mathcal{F}\mathcal{M}'}=0.622$, and $\widehat{p}_{\mathcal{F}\mathcal{F}'}=0.595$. Similarly we obtain the estimates for the other probabilities w.r.t. observation bias: $\widehat{q}_{\mathcal{M}\mathcal{M}'}=0.053,\ \widehat{q}_{\mathcal{M}\mathcal{F}'}=0.037,\  \widehat{q}_{\mathcal{F}\mathcal{M}'}=0.037$, and $\widehat{q}_{\mathcal{F}\mathcal{F}'}=0.046$.
    \item \emph{Last FM 360K}: The associated task is to predict whether the user likes the artist or not. This dataset contains 359,347 users, 294,015 artists, and 17,559,530 play counts, i.e., rating matrix is 0.02\% full.\footnote{http://ocelma.net/MusicRecommendationDataset/lastfm-360K.html} The data for play counts is converted to binary rating: $+1$ if counts $>$ average, otherwise $-1$. We divide user and item groups based on gender and genre, respectively. Since this dataset only contains gender information, we use Last.fm API\footnote{http://www.last.fm/api} to collect the genre of corresponding artist's music; the tag was associated with 5,706 artists. 
    % We also randomly select 5000 male and 5000 female users. Among 10 genres, we choose hip-hop and musical for male and female preferred genres, respectively. The final rating matrix of 10,000 users and 5,706 artists is 0.55\% full. 
    % From the real data, we obtain empirical estimates for the interested probabilities w.r.t. population imbalance: $\widehat{p}_{\mathcal{M}\mathcal{M}'}=0.548,\ \widehat{p}_{\mathcal{M}\mathcal{F}'}=0.421,\  \widehat{p}_{\mathcal{F}\mathcal{M}'}=0.438$ and $\widehat{p}_{\mathcal{F}\mathcal{F}'}=0.529$. Similarly we obtain the estimates for the other probabilities w.r.t. observation bias: $\widehat{q}_{\mathcal{M}\mathcal{M}'}=0.0054,\ \widehat{q}_{\mathcal{M}\mathcal{F}'}=0.0011,\  \widehat{q}_{\mathcal{F}\mathcal{M}'}=0.0036$ and $\widehat{q}_{\mathcal{F}\mathcal{F}'}=0.0038$.
\end{itemize}

We run experiments employing both matrix factorization (MF) based and autoencoder (AE) based techniques.
In Appendix~\ref{sec: appendix_real}, we leave a more detailed explanation of experiments on the real datasets and the results of AE-based technique. The results of real data experiments are listed in Tables~\ref{table:movielens_MF} and~\ref{table:LastFM_MF}. As in the synthetic data setting, we can make two relevant observations. All baseline algorithms fail to respect our metric (${\sf DEE}$) while meeting their own. We also see that our algorithm exhibits significant performances for all the fairness measures except for ${\sf VAL}$ which does not have close relationship with the equal experience that we aim at. 

\begin{table*}[h!]
  \caption{Prediction error (RMSE) and fairness performances of the matrix factorization based algorithm on \emph{MovieLens 1M dataset}. The boldface indicates the best result and the underline denotes the second best. Our algorithm offers great fairness performances for all the measures, except for ${\sf VAL}$, which our framework does not target.}
  \label{table:movielens_MF}
  \centering
  \resizebox{6in}{!}{
  \begin{tabular}{cccccc}
    \toprule
    Measure & RMSE & $\sf DEE$ & $\sf VAL$ & $\sf UGF$ & $\sf CVS$ \\
    \midrule
    Unfair   & 0.8541 $\pm$ 0.0033     & 0.2447 $\pm$ 0.0134 & 0.3227 $\pm$ 0.0031    & 0.0058 $\pm$ 0.0042   & 0.1291 $\pm$ 0.0079  \\
    Ours (\sf{DEE})   & 0.8641 $\pm$ 0.0047     & \textbf{0.0014} $\pm$ \textbf{0.0008}   & 0.2941 $\pm$ 0.0024    & \underline{0.0018 $\pm$ 0.0016}  & \underline{0.0007 $\pm$ 0.0004} \\
    Ours (\sf{DER})   & 0.8526 $\pm$ 0.0029     & {0.0114} $\pm$ {0.0041}   & 0.3332 $\pm$ 0.0050    & {0.0055 $\pm$ 0.0022}  & {0.0014 $\pm$ 0.0001} \\
    $\sf VAL$-based   & 0.8529 $\pm$ 0.0011    & 0.3659 $\pm$ 0.0033     & \textbf{0.0942} $\pm$ \textbf{0.0016}  & 0.0261 $\pm$ 0.0020 & 0.1388 $\pm$ 0.0030\\
    $\sf UGF$-based   & 0.8550 $\pm$ 0.0015     & 0.2492 $\pm$ 0.0100  & 0.3285 $\pm$ 0.0051  & \textbf{0.0001} $\pm$ \textbf{0.0001} & 0.1355 $\pm$ 0.0038\\
    $\sf CVS$-based   & 0.8549 $\pm$ 0.0018     & 0.0721 $\pm$ 0.0069    & 0.3319 $\pm$ 0.0046  & 0.0065 $\pm$ 0.0042 & \textbf{0.0002} $\pm$ \textbf{3.45e-5}\\
    \bottomrule
  \end{tabular}}
\end{table*}

\begin{table*}[h!]
  \caption{Prediction error (RMSE) and fairness performances of the matrix factorization based algorithm on \emph{Last FM 360K dataset}.}
  \label{table:LastFM_MF}
  \centering
  \resizebox{6in}{!}{\begin{tabular}{cccccc}
    \toprule
    Measure & RMSE & $\sf DEE$ & $\sf VAL$  & $\sf UGF$ & $\sf CVS$ \\
    \midrule
    Unfair   & 0.6720 $\pm$ 0.0024     & 0.0840 $\pm$ 0.0110 & 0.2297 $\pm$ 0.0020    & 0.0404 $\pm$ 0.0033   & 0.0204 $\pm$ 0.0104  \\
    Ours (\sf{DEE})   & 0.6892 $\pm$ 0.0040     & \textbf{0.0082} $\pm$ \textbf{0.0023} & 0.2777 $\pm$ 0.0048    & \underline{0.0161 $\pm$ 0.0009}  & \underline{0.0040 $\pm$ 0.0011} \\
    Ours (\sf{DER})  & 0.6830 $\pm$ 0.0033  & {0.0711} $\pm$ {0.0140}   & 0.2588 $\pm$ 0.0024    & {0.0356 $\pm$ 0.0072}  & {0.0047 $\pm$ 0.0007} \\
    $\sf VAL$-based   & 0.6802 $\pm$ 0.0006     & 0.1461 $\pm$ 0.0216  & \textbf{0.0016} $\pm$ \textbf{2.03e-5}  & 0.0234 $\pm$ 0.0020 & 0.0324 $\pm$ 0.0202\\
    $\sf UGF$-based   & 0.6705 $\pm$ 0.0030     & 0.0644 $\pm$ 0.0313  & 0.2366 $\pm$ 0.0016  & \textbf{0.0011} $\pm$ \textbf{4.03e-5} & 0.3221 $\pm$ 0.0157\\
    $\sf CVS$-based  & 0.6758 $\pm$ 0.0037     & 0.0791 $\pm$ 0.0136  & 0.2448 $\pm$ 0.0034  & 0.0373 $\pm$ 0.0063 & \textbf{0.0012} $\pm$ \textbf{0.0003}\\
    \bottomrule
  \end{tabular}}
\end{table*}

\section{Extension}
\label{sec: extension}
In this section, we discuss the extension of our work: (i) introducing a new notion that bears a similar spirit to \emph{equalized odds}, and (ii) applying our notion to the \emph{fair ranking} context~\cite{zehlike2017}. 

Our fairness notion \textit{equal experience} aims at recommending a variety of items for all user groups. \emph{Demographic parity} is similar to our notion in the sense of considering the irrelevancy of predictions to groups. On the other hand, \emph{equalized odds} is the fairness notion that encourages equal error rates (e.g., true/negative positive rate) across user groups by employing the ground-truth label $Y$. Similar to \emph{equalized odds}, our measure in recommender systems can readily be extended to a setting in which the ground-truth label is available to exploit. The key idea is to promote $\widetilde{Y}\perp (Z_{\sf user}, Z_{\sf item})|Y$. We can employ the proposed optimization framework via a new measure to promote the notion. See Appendix~\ref{sec: extended fairness notion} for details.

Many end-to-end recommender systems offer a recommendation list via two processes: (i) candidate generation and (ii) ranking. In this work, we focus on the first candidate generation for which we built collaborative filtering. But our proposed notion can also be applicable in generating an end ranked list. The idea behind the end-ranked list generation is to define an indicator function, say $R$, which returns 1 when the item of interest belongs to, say top-$K$ item set (0 otherwise). In this case, a similar notion of the same structure $R\perp (Z_{\sf user}, Z_{\sf item})$ serves a proper role. We also provide experimental results which demonstrate the effectiveness of the proposed method in the top-$K$ recommendation setting (See Appendix~\ref{sec: appendix_recommendation}). 

% We leave detailed implementation of this notion under the fair ranking context for a future work.

\section{Conclusion}
\label{sec: conclusion}
We introduced a novel fairness notion, \emph{equal experience}, capable of respecting the desired requirements for fair recommender systems: independence between preference predictions and user groups; conditional independence for a certain user group, and vice versa for item groups. The notion also seamlessly integrates into prior fairness algorithms. Extensive experiments revealed the existence of unfairness (or bias) w.r.t. \emph{equal experience}, and our fair optimization framework successfully mitigates such unfairness with minimal degradation in prediction accuracy. Our future work of interest is four-folded: (i) merging our notion with unexamined algorithms relying upon R\'{e}nyi correlation or Wasserstein distance; (ii) constructing a \emph{robust} and fair recommender system in the presence of data poisoning; and (iii) developing a blind fair recommender system without sensitive attributes. 

% and (iv) extending our notion to fair ranking. 

\nocite{langley00}

\bibliography{reference}

\begin{thebibliography}{53}
\providecommand{\natexlab}[1]{#1}
\providecommand{\url}[1]{\texttt{#1}}
\expandafter\ifx\csname urlstyle\endcsname\relax
  \providecommand{\doi}[1]{doi: #1}\else
  \providecommand{\doi}{doi: \begingroup \urlstyle{rm}\Url}\fi

\bibitem[Abadir \& Magnus(2005)Abadir and Magnus]{abadir2005}
Abadir, K.~M. and Magnus, J.~R.
\newblock \emph{Matrix algebra}, volume~1.
\newblock Cambridge University Press, 2005.

\bibitem[Beutel et~al.(2019)Beutel, Chen, Doshi, Qian, Wei, Wu, Heldt, Zhao,
  Hong, Chi, et~al.]{beutel2019}
Beutel, A., Chen, J., Doshi, T., Qian, H., Wei, L., Wu, Y., Heldt, L., Zhao,
  Z., Hong, L., Chi, E.~H., et~al.
\newblock Fairness in recommendation ranking through pairwise comparisons.
\newblock In \emph{Proceedings of the 25th ACM SIGKDD International Conference
  on Knowledge Discovery \& Data Mining}, pp.\  2212--2220, 2019.

\bibitem[Biega et~al.(2018)Biega, Gummadi, and Weikum]{biega2018}
Biega, A.~J., Gummadi, K.~P., and Weikum, G.
\newblock Equity of attention: Amortizing individual fairness in rankings.
\newblock In \emph{The 41st international acm sigir conference on research \&
  development in information retrieval}, pp.\  405--414, 2018.

\bibitem[Burke(2017)]{burke2017}
Burke, R.
\newblock Multisided fairness for recommendation.
\newblock \emph{arXiv preprint arXiv:1707.00093}, 2017.

\bibitem[Cand{\`e}s \& Recht(2009)Cand{\`e}s and Recht]{candes2009}
Cand{\`e}s, E.~J. and Recht, B.
\newblock Exact matrix completion via convex optimization.
\newblock \emph{Foundations of Computational mathematics}, 9\penalty0
  (6):\penalty0 717--772, 2009.

\bibitem[Celma(2010)]{data_lastfm}
Celma, O.
\newblock \emph{Music Recommendation and Discovery in the Long Tail}.
\newblock Springer, 2010.

\bibitem[Cho et~al.(2020{\natexlab{a}})Cho, Hwang, and Suh]{cho2020kde}
Cho, J., Hwang, G., and Suh, C.
\newblock A fair classifier using kernel density estimation.
\newblock \emph{Advances in Neural Information Processing Systems},
  33:\penalty0 15088--15099, 2020{\natexlab{a}}.

\bibitem[Cho et~al.(2020{\natexlab{b}})Cho, Hwang, and Suh]{cho2020mi}
Cho, J., Hwang, G., and Suh, C.
\newblock A fair classifier using mutual information.
\newblock In \emph{2020 IEEE International Symposium on Information Theory
  (ISIT)}, pp.\  2521--2526. IEEE, 2020{\natexlab{b}}.

\bibitem[Cover(1999)]{cover1999}
Cover, T.~M.
\newblock \emph{Elements of information theory}.
\newblock John Wiley \& Sons, 1999.

\bibitem[Davis et~al.(2011)Davis, Lii, and Politis]{davis2011}
Davis, R.~A., Lii, K.-S., and Politis, D.~N.
\newblock Remarks on some nonparametric estimates of a density function.
\newblock In \emph{Selected Works of Murray Rosenblatt}, pp.\  95--100.
  Springer, 2011.

\bibitem[Donini et~al.(2018)Donini, Oneto, Ben-David, Shawe-Taylor, and
  Pontil]{donini2018}
Donini, M., Oneto, L., Ben-David, S., Shawe-Taylor, J.~S., and Pontil, M.
\newblock Empirical risk minimization under fairness constraints.
\newblock In \emph{Advances in Neural Information Processing Systems 31
  (NeurIPS)}, 2018.

\bibitem[Dwork et~al.(2012)Dwork, Hardt, Pitassi, Reingold, and
  Zemel]{dwork2012}
Dwork, C., Hardt, M., Pitassi, T., Reingold, O., and Zemel, R.~S.
\newblock Fairness through awareness.
\newblock \emph{Innovations in Theoretical Computer Science Conferennce
  (ITCS)}, 2012.

\bibitem[Fazel(2002)]{fazel2002}
Fazel, M.
\newblock \emph{Matrix rank minimization with applications}.
\newblock PhD thesis, PhD thesis, Stanford University, 2002.

\bibitem[Feldman et~al.(2015)Feldman, Friedler, Moeller, Scheidegger, and
  Venkatasubramanian]{feldman2015}
Feldman, M., Friedler, S.~A., Moeller, J., Scheidegger, C., and
  Venkatasubramanian, S.
\newblock Certifying and removing disparate impact.
\newblock In \emph{proceedings of the 21th ACM SIGKDD international conference
  on knowledge discovery and data mining}, pp.\  259--268, 2015.

\bibitem[Garg et~al.(2018)Garg, Perot, Limtiaco, Taly, hsin Chi, and
  Beutel]{garg2018}
Garg, S., Perot, V., Limtiaco, N., Taly, A., hsin Chi, E.~H., and Beutel, A.
\newblock Counterfactual fairness in text classification through robustness.
\newblock \emph{AAAI/ACM Conference on Artificial Intelligence, Ethics, and
  Society (AIES)}, 2018.

\bibitem[G\'{e}ron(2017)]{Geron}
G\'{e}ron, A.
\newblock \emph{Hands-On Machine Learning with Scikit-Learn \& TensorFlow}.
\newblock O'Reilly, 2017.

\bibitem[G{\'e}ron(2019)]{geron2019}
G{\'e}ron, A.
\newblock \emph{Hands-on machine learning with Scikit-Learn, Keras, and
  TensorFlow: Concepts, tools, and techniques to build intelligent systems}.
\newblock O'Reilly Media, 2019.

\bibitem[Hardt et~al.(2016)Hardt, Price, Price, and Srebro]{hardt2016}
Hardt, M., Price, E., Price, E., and Srebro, N.
\newblock Equality of opportunity in supervised learning.
\newblock In \emph{Advances in Neural Information Processing Systems 29
  (NeurIPS)}, 2016.

\bibitem[Harper \& Konstan(2015)Harper and Konstan]{data_movielens}
Harper, F.~M. and Konstan, J.~A.
\newblock The movielens datasets: History and context.
\newblock \emph{ACM Transactions on Interactive Intelligent Systems (TiiS)},
  2015.

\bibitem[He et~al.(2017)He, Liao, Zhang, Nie, Hu, and Chua]{he2017}
He, X., Liao, L., Zhang, H., Nie, L., Hu, X., and Chua, T.-S.
\newblock Neural collaborative filtering.
\newblock In \emph{Proceedings of the 26th international conference on world
  wide web}, pp.\  173--182, 2017.

\bibitem[Huber(1992)]{huber1992}
Huber, P.~J.
\newblock Robust estimation of a location parameter.
\newblock In \emph{Breakthroughs in statistics}, pp.\  492--518. Springer,
  1992.

\bibitem[Jiang et~al.(2019)Jiang, Pacchiano, Stepleton, Jiang, and
  Chiappa]{jiang2019}
Jiang, R., Pacchiano, A., Stepleton, T., Jiang, H., and Chiappa, S.
\newblock Wasserstein fair classification.
\newblock In \emph{Proceedings of the Thirty-Fifth Conference on Uncertainty in
  Artificial Intelligence, (UAI)}, 2019.

\bibitem[Jiang et~al.(2020)Jiang, Pacchiano, Stepleton, Jiang, and
  Chiappa]{jiang2020}
Jiang, R., Pacchiano, A., Stepleton, T., Jiang, H., and Chiappa, S.
\newblock Wasserstein fair classification.
\newblock In \emph{Uncertainty in Artificial Intelligence}, pp.\  862--872.
  PMLR, 2020.

\bibitem[Kamishima \& Akaho(2017)Kamishima and Akaho]{kamishima2017}
Kamishima, T. and Akaho, S.
\newblock Considerations on recommendation independence for a find-good-items
  task.
\newblock 2017.

\bibitem[Kamishima et~al.(2012)Kamishima, Akaho, Asoh, and
  Sakuma]{kamishima2012}
Kamishima, T., Akaho, S., Asoh, H., and Sakuma, J.
\newblock Fairness-aware classifier with prejudice remover regularizer.
\newblock In \emph{Joint European Conference on Machine Learning and Knowledge
  Discovery in Databases}, pp.\  35--50. Springer, 2012.

\bibitem[Kingma \& Ba(2014{\natexlab{a}})Kingma and Ba]{ref20_adam}
Kingma, D. and Ba, J.
\newblock Adam: {A} method for stochastic optimization.
\newblock \emph{arXiv preprint arXiv:1412.6980}, 2014{\natexlab{a}}.

\bibitem[Kingma \& Ba(2014{\natexlab{b}})Kingma and Ba]{kingma2014}
Kingma, D.~P. and Ba, J.
\newblock Adam: A method for stochastic optimization.
\newblock \emph{arXiv preprint arXiv:1412.6980}, 2014{\natexlab{b}}.

\bibitem[Koren(2008)]{koren2008}
Koren, Y.
\newblock Factorization meets the neighborhood: a multifaceted collaborative
  filtering model.
\newblock In \emph{Proceedings of the 14th ACM SIGKDD international conference
  on Knowledge discovery and data mining}, pp.\  426--434, 2008.

\bibitem[Koren et~al.(2009)Koren, Bell, and Volinsky]{koren2009}
Koren, Y., Bell, R., and Volinsky, C.
\newblock Matrix factorization techniques for recommender systems.
\newblock \emph{Computer}, 42\penalty0 (8):\penalty0 30--37, 2009.

\bibitem[Kusner et~al.(2017)Kusner, Loftus, Russell, and Silva]{kusner2017}
Kusner, M.~J., Loftus, J., Russell, C., and Silva, R.
\newblock Counterfactual fairness.
\newblock In \emph{Advances in Neural Information Processing Systems 30
  (NeurIPS)}, 2017.

\bibitem[Lee et~al.(2018)Lee, Lee, and Suh]{lee2018}
Lee, K., Lee, Y.~H., and Suh, C.
\newblock Alternating autoencoders for matrix completion.
\newblock In \emph{2018 IEEE Data Science Workshop (DSW)}, pp.\  130--134.
  IEEE, 2018.

\bibitem[Li et~al.(2021)Li, Chen, Fu, Ge, and Zhang]{li2021}
Li, Y., Chen, H., Fu, Z., Ge, Y., and Zhang, Y.
\newblock User-oriented fairness in recommendation.
\newblock \emph{arXiv preprint arXiv:2104.10671}, 2021.

\bibitem[Mary et~al.(2019)Mary, Calauzenes, and El~Karoui]{mary2019}
Mary, J., Calauzenes, C., and El~Karoui, N.
\newblock Fairness-aware learning for continuous attributes and treatments.
\newblock In \emph{International Conference on Machine Learning}, pp.\
  4382--4391. PMLR, 2019.

\bibitem[Mehrotra et~al.(2018)Mehrotra, McInerney, Bouchard, Lalmas, and
  Diaz]{mehrotra2018}
Mehrotra, R., McInerney, J., Bouchard, H., Lalmas, M., and Diaz, F.
\newblock Towards a fair marketplace: Counterfactual evaluation of the
  trade-off between relevance, fairness \& satisfaction in recommendation
  systems.
\newblock In \emph{Proceedings of the 27th acm international conference on
  information and knowledge management}, pp.\  2243--2251, 2018.

\bibitem[Nabi \& Shpitser(2018)Nabi and Shpitser]{nabi2018}
Nabi, R. and Shpitser, I.
\newblock Fair inference on outcomes.
\newblock In \emph{Proceedings of the Thirty-Second {AAAI} Conference on
  Artificial Intelligence (AAAI)}, 2018.

\bibitem[Narasimhan et~al.(2020)Narasimhan, Cotter, Gupta, and
  Wang]{narasimhan2020}
Narasimhan, H., Cotter, A., Gupta, M., and Wang, S.
\newblock Pairwise fairness for ranking and regression.
\newblock In \emph{Proceedings of the AAAI Conference on Artificial
  Intelligence}, volume~34, pp.\  5248--5255, 2020.

\bibitem[Paszke et~al.(2019)Paszke, Gross, Massa, Lerer, Bradbury, Chanan,
  Killeen, Lin, Gimelshein, Antiga, Desmaison, Kopf, Yang, DeVito, Raison,
  Tejani, Chilamkurthy, Steiner, Fang, Bai, and Chintala]{pytorch}
Paszke, A., Gross, S., Massa, F., Lerer, A., Bradbury, J., Chanan, G., Killeen,
  T., Lin, Z., Gimelshein, N., Antiga, L., Desmaison, A., Kopf, A., Yang, E.,
  DeVito, Z., Raison, M., Tejani, A., Chilamkurthy, S., Steiner, B., Fang, L.,
  Bai, J., and Chintala, S.
\newblock Pytorch: An imperative style, high-performance deep learning library.
\newblock \emph{Advances in Neural Information Processing Systems 32
  (NeurIPS)}, 2019.

\bibitem[Robbins \& Monro(1951)Robbins and Monro]{robbins1951}
Robbins, H. and Monro, S.
\newblock A stochastic approximation method.
\newblock \emph{The annals of mathematical statistics}, pp.\  400--407, 1951.

\bibitem[Russell et~al.(2017)Russell, Kusner, Loftus, and Silva]{russell2017}
Russell, C., Kusner, M.~J., Loftus, J., and Silva, R.
\newblock When worlds collide: Integrating different counterfactual assumptions
  in fairness.
\newblock In \emph{Advances in Neural Information Processing Systems 30
  (NeurIPS)}, 2017.

\bibitem[Salakhutdinov et~al.(2007)Salakhutdinov, Mnih, and
  Hinton]{salakhutdinov2007}
Salakhutdinov, R., Mnih, A., and Hinton, G.
\newblock Restricted boltzmann machines for collaborative filtering.
\newblock In \emph{Proceedings of the 24th international conference on Machine
  learning}, pp.\  791--798, 2007.

\bibitem[Schnabel et~al.(2016)Schnabel, Swaminathan, Singh, Chandak, and
  Joachims]{schnabel2016}
Schnabel, T., Swaminathan, A., Singh, A., Chandak, N., and Joachims, T.
\newblock Recommendations as treatments: Debiasing learning and evaluation.
\newblock In \emph{International Conference on Machine Learning}, pp.\
  1670--1679. PMLR, 2016.

\bibitem[Sedhain et~al.(2015)Sedhain, Menon, Sanner, and Xie]{sedhain2015}
Sedhain, S., Menon, A.~K., Sanner, S., and Xie, L.
\newblock Autorec: Autoencoders meet collaborative filtering.
\newblock In \emph{Proceedings of the 24th international conference on World
  Wide Web}, pp.\  111--112, 2015.

\bibitem[Singh \& Joachims(2018)Singh and Joachims]{singh2018}
Singh, A. and Joachims, T.
\newblock Fairness of exposure in rankings.
\newblock In \emph{Proceedings of the 24th ACM SIGKDD International Conference
  on Knowledge Discovery \& Data Mining}, pp.\  2219--2228, 2018.

\bibitem[Woodworth et~al.(2017)Woodworth, Gunasekar, Ohannessian, and
  Srebro]{woodworth2017}
Woodworth, B., Gunasekar, S., Ohannessian, M.~I., and Srebro, N.
\newblock Learning non-discriminatory predictors.
\newblock In \emph{Conference on Learning Theory}, pp.\  1920--1953. PMLR,
  2017.

\bibitem[Wu et~al.(2019)Wu, Zhang, and Wu]{wu2019}
Wu, Y., Zhang, L., and Wu, X.
\newblock Counterfactual fairness: Unidentification, bound and algorithm.
\newblock In \emph{Proceedings of the Twenty-Eighth International Joint
  Conference on Artificial Intelligence, {IJCAI-19}}, 2019.

\bibitem[Xiao et~al.(2017)Xiao, Min, Yongfeng, Zhaoquan, Yiqun, and
  Shaoping]{xiao2017}
Xiao, L., Min, Z., Yongfeng, Z., Zhaoquan, G., Yiqun, L., and Shaoping, M.
\newblock Fairness-aware group recommendation with pareto-efficiency.
\newblock In \emph{Proceedings of the Eleventh ACM Conference on Recommender
  Systems}, pp.\  107--115, 2017.

\bibitem[Yao \& Huang(2017)Yao and Huang]{yao2017}
Yao, S. and Huang, B.
\newblock Beyond parity: Fairness objectives for collaborative filtering.
\newblock \emph{arXiv preprint arXiv:1705.08804}, 2017.

\bibitem[Zafar et~al.(2017{\natexlab{a}})Zafar, Valera, Gomez~Rodriguez, and
  Gummadi]{zafar2017a}
Zafar, M.~B., Valera, I., Gomez~Rodriguez, M., and Gummadi, K.~P.
\newblock Fairness beyond disparate treatment \& disparate impact: Learning
  classification without disparate mistreatment.
\newblock In \emph{Proceedings of the 26th international conference on world
  wide web}, pp.\  1171--1180, 2017{\natexlab{a}}.

\bibitem[Zafar et~al.(2017{\natexlab{b}})Zafar, Valera, Rogriguez, and
  Gummadi]{zafar2017b}
Zafar, M.~B., Valera, I., Rogriguez, M.~G., and Gummadi, K.~P.
\newblock Fairness constraints: Mechanisms for fair classification.
\newblock In \emph{Artificial Intelligence and Statistics}, pp.\  962--970.
  PMLR, 2017{\natexlab{b}}.

\bibitem[Zehlike et~al.(2017)Zehlike, Bonchi, Castillo, Hajian, Megahed, and
  Baeza-Yates]{zehlike2017}
Zehlike, M., Bonchi, F., Castillo, C., Hajian, S., Megahed, M., and
  Baeza-Yates, R.
\newblock Fa* ir: A fair top-k ranking algorithm.
\newblock In \emph{Proceedings of the 2017 ACM on Conference on Information and
  Knowledge Management}, pp.\  1569--1578, 2017.

\bibitem[Zhang et~al.(2018)Zhang, Lemoine, and Mitchell]{zhang2018mitigating}
Zhang, B.~H., Lemoine, B., and Mitchell, M.
\newblock Mitigating unwanted biases with adversarial learning.
\newblock In \emph{Proceedings of the 2018 AAAI/ACM Conference on AI, Ethics,
  and Society}, pp.\  335--340, 2018.

\bibitem[Zhang \& Bareinboim(2018{\natexlab{a}})Zhang and
  Bareinboim]{zhang2018decision}
Zhang, J. and Bareinboim, E.
\newblock Fairness in decision-making — the causal explanation formula.
\newblock In \emph{Proceedings of the Thirty-Second {AAAI} Conference on
  Artificial Intelligence (AAAI)}, 2018{\natexlab{a}}.

\bibitem[Zhang \& Bareinboim(2018{\natexlab{b}})Zhang and
  Bareinboim]{zhang2018eo}
Zhang, J. and Bareinboim, E.
\newblock Equality of opportunity in classification: A causal approach.
\newblock In \emph{Advances in Neural Information Processing Systems 31
  (NeurIPS)}, 2018{\natexlab{b}}.

\end{thebibliography}
\bibliographystyle{icml2022}

%%%%%%%%%%%%%%%%%%%%%%%%%%%%%%%%%%%%%%%%%%%%%%%%%%%%%%%%%%%%%%%%%%%%%%%%%%%%%%%
%%%%%%%%%%%%%%%%%%%%%%%%%%%%%%%%%%%%%%%%%%%%%%%%%%%%%%%%%%%%%%%%%%%%%%%%%%%%%%%
% APPENDIX
%%%%%%%%%%%%%%%%%%%%%%%%%%%%%%%%%%%%%%%%%%%%%%%%%%%%%%%%%%%%%%%%%%%%%%%%%%%%%%%
%%%%%%%%%%%%%%%%%%%%%%%%%%%%%%%%%%%%%%%%%%%%%%%%%%%%%%%%%%%%%%%%%%%%%%%%%%%%%%%
\newpage
\appendix
\onecolumn

\section{Appendix}
\subsection{Outline}
\label{sec: appendix_outline}
We first provide further explanation as to how to implement the KDE technique. Next, we provide a detailed description of experimental settings on both synthetic and real datasets: MovieLens 1M~\citep{data_movielens} and Last FM 360K~\citep{data_lastfm}. We then present additional experimental results which are not included in the main paper due to space limitation, as well as provide a complexity analysis of our approach. We provide a detailed explanation for the extension of our measure to the fairness notion: $\widetilde{Y}\perp (Z_{\sf user}, Z_{\sf item})|Y$, and present experimental results for this extension. We also provide elaboration on the extension of our measure to top-$K$ recommendation setting, and present experimental results on this setting.      

\subsection{Implementation of the KDE technique}
\label{sec:appendix_kde}
In order to approximate $\nabla_w\textsf{DEE}$, we first calculate the gradient of $ \widehat{\mathbb{P}}(\widetilde{Y} = 1)$ employing the technique in~\cite{cho2020kde}:
\begin{align}
\label{eq:DP_grad}
    \nabla_w\widehat{\mathbb{P}}(\widetilde{Y}=1)=
    \frac{1}{nmh}\sum_{i=1}^n\sum_{j=1}^mf_k\left({\frac{\tau-\widehat{M}_{ij}}{h}}\right)\cdot\nabla_w{\widehat{M}_{ij}}.
\end{align}
We can apply the same procedures w.r.t. the second interested probability  $\widehat{\mathbb{P}}(\widetilde{Y} =1 | Z_{\sf item} = z_2, Z_{\sf user} = z_1)$. 
Merging~\eqref{eq:DP_grad} and the counterpart w.r.t. the second probability, we can readily obtain:
\begin{align}
\begin{split}
    \nabla_w\textsf{DEE}\approx
    \sum_{z_1 \in \mathcal{Z}_{\sf user} } & \sum_{z_2 \in {\cal Z}_{\sf item} }   H'_{\delta}\left( \widehat{\mathbb{P}}(\widetilde{Y} =1 | Z_{\sf item} = z_2, Z_{\sf user} = z_1) -\widehat{\mathbb{P}}(\widetilde{Y}=1)  \right) \\
  & \qquad \times \nabla_w \left( \widehat{\mathbb{P}}(\widetilde{Y} =1 | Z_{\sf item} = z_2, Z_{\sf user} = z_1)-\widehat{\mathbb{P}}(\widetilde{Y}=1)  \right)
\end{split}
\end{align}
where $H_{\delta}(x)$ denotes the Huber loss~\citep{huber1992} that takes $\frac{1}{2}x^2 $ when  $|x|\leq\delta$; otherwise $\delta(|x|-\frac{1}{2}\delta)$. 

\subsection{Synthetic dataset experiments}
\label{sec: appendix_synthetic}

\begin{figure}[h]
    \centering  
    \includegraphics[height=4.5cm]{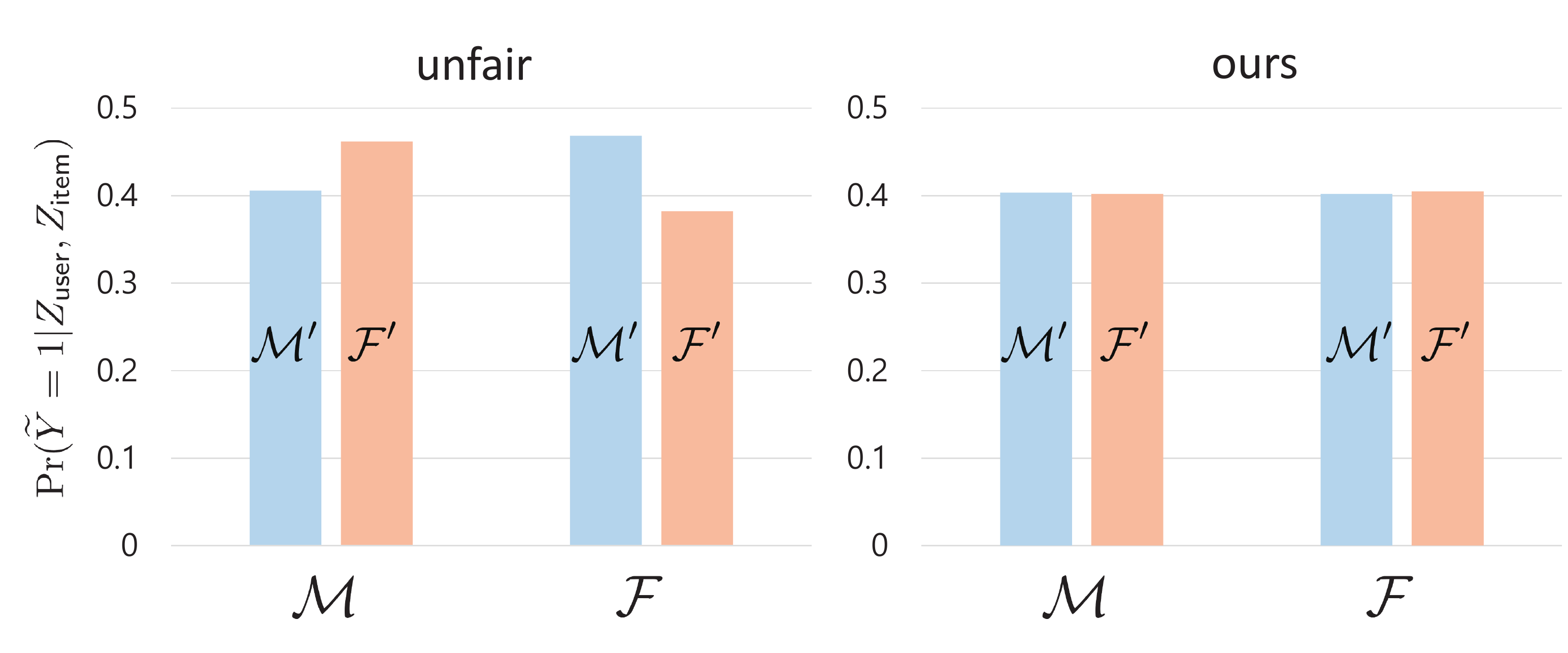}
    \caption{{Predicted preference rate of item groups for every user group $\Pr(\tilde{Y}=1|Z_{\sf user}, Z_{\sf item})$ on the synthetic dataset. The dataset preserves observation bias $(q_0,q_1)=(0.2, 0.01)$ while exhibiting no population imbalance $(p_0,p_1)=(0.4, 0.4)$.}}
    \label{fig:3}
\end{figure}

\begin{figure}[h]
    \centering  
    \includegraphics[height=4.5cm]{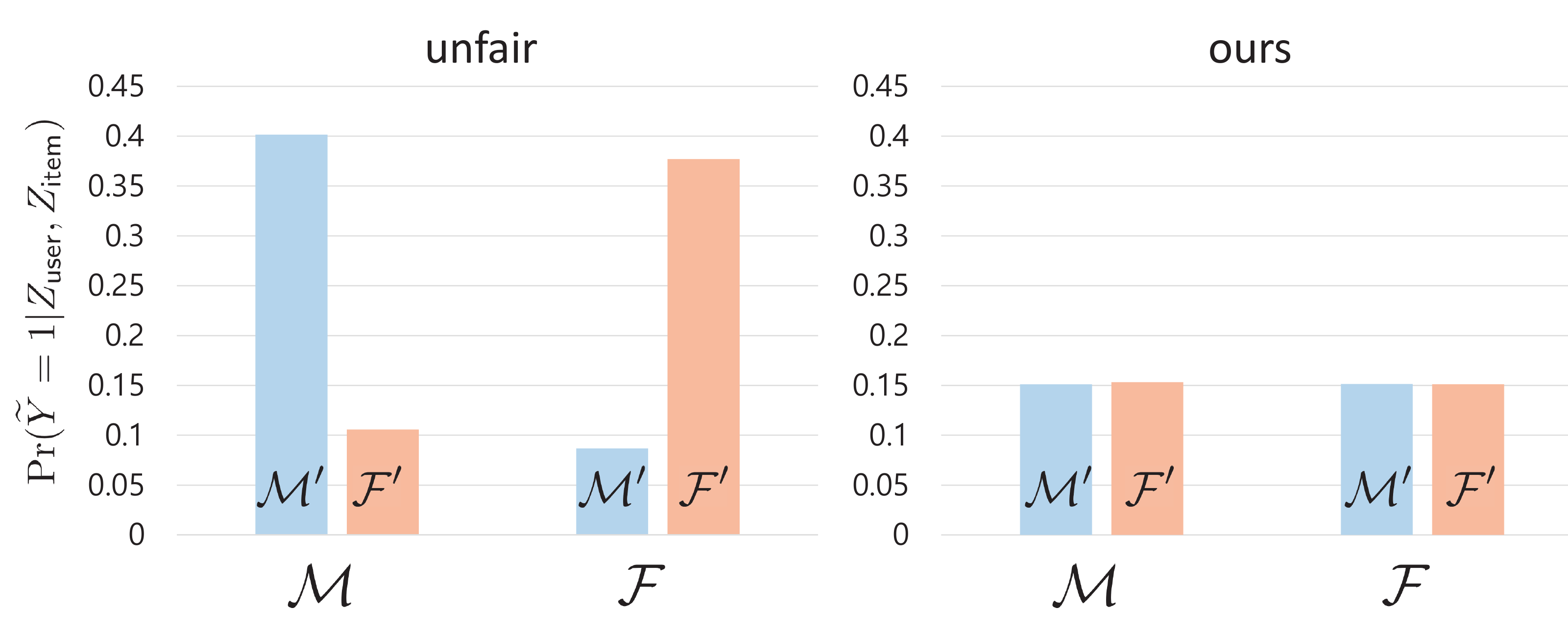}
    \caption{{Predicted preference rate of item groups for every user group $\Pr(\tilde{Y}=1|Z_{\sf user}, Z_{\sf item})$ on the synthetic dataset. The dataset preserves population imbalance $(p_0,p_1)=(0.4, 0.1)$ while exhibiting no observation bias $(q_0,q_1)=(0.2, 0.2)$.}}
    \label{fig:4}
\end{figure}

We consider a setting in which $(r, n, m) = (20, 600, 400)$. The synthetic data generated under the setting is randomly split into two subsets: 90\% train set and 10\% test set. Since $M_{ij} \in \{+1,-1\}$, we set the threshold $\tau=0$, i.e., $\widetilde{Y}={\bf 1} \{ \widehat{Y} \geq 0 \}$. We train a matrix factorization (MF) based recommender system with the same rank as that of the dataset, i.e., $L\in \mathbb{R}^{600 \times 20}$ and $R\in \mathbb{R}^{20 \times 400}$. We set hyperparameters $(\delta,h)=(0.01, 0.01)$ and $\lambda=0.99$ for KDE-based algorithm implementation. We use Adam optimizer for 1,000 iterations using full gradient. The learning rate is set to 1e-3 and $(\beta_1, \beta_2)=(0.9, 0.999)$. The additional experimental results in a variety of scenarios are listed from Table~\ref{table:synthetic_MF1} to~\ref{table:synthetic_MF5}. We also visualize how the predicted preference rate of item groups for every user group $\Pr(\tilde{Y}=1|Z_{\sf user}, Z_{\sf item})$ changes under our framework. See Fig.~\ref{fig:3} and~\ref{fig:4}. 

%scaling done
\begin{table}[h]
  \caption{Prediction error (RMSE) and fairness performances on the synthetic dataset. The dataset preserves observation bias $(q_0,q_1)=(0.2, 0.02)$ while exhibiting no population imbalance $(p_0,p_1)=(0.4, 0.4)$. The boldface indicates the best result and the underline denotes the second best. Each baseline, say ${\sf VAL}$-based approach, enjoys the best fairness performance only for the measure focused therein, ${\sf VAL}$, while it does not work well under the other fairness measures. On the other hand, our algorithm offers great fairness performances for all the measures, except for ${\sf VAL}$, which our framework does not target.}
  \label{table:synthetic_MF1}
  \centering
  \resizebox{\textwidth}{!}{\begin{tabular}{cccccc}
    \toprule
    Measure & RMSE & $\sf DEE$ & $\sf VAL$ & $\sf UGF$ & $\sf CVS$ \\
    \midrule
    Unfair   & 0.8423 $\pm$ 0.0086 & 0.1002 $\pm$ 0.0264 & 0.2992 $\pm$ 0.0142 & 0.0160 $\pm$ 0.0031 & 0.0066 $\pm$ 0.0034 \\
    Ours (\sf{DEE})   & 0.8611 $\pm$ 0.0116 & \textbf{0.0022} $\pm$ \textbf{0.0003} & 0.3273 $\pm$ 0.0043 & \underline{0.0006 $\pm$ 0.0002}& \underline{0.0003 $\pm$ 0.0001}\\
    Ours (\sf{DER})   & 0.8605 $\pm$ 0.0069 & 0.0414 $\pm$ 0.0059 & {0.3130} $\pm$ {0.0103} & 0.0207 $\pm$ 0.0030 & 0.0001 $\pm$ 0.0001\\
    $\sf VAL$-based   & 0.8460 $\pm$ 0.0081 & 0.0829 $\pm$ 0.0035 & \textbf{0.0003} $\pm$ \textbf{1.25e-5} & 0.0138 $\pm$ 0.0019 & 0.0066 $\pm$ 0.0014\\
    $\sf UGF$-based   & 0.8546 $\pm$ 0.0050 & 0.1137 $\pm$ 0.0149 & 0.3163 $\pm$ 0.0036 & \textbf{0.0004} $\pm$ \textbf{0.0001} & 0.0195 $\pm$ 0.0031 \\
    $\sf CVS$-based  & 0.8622 $\pm$ 0.0072 & 0.1189 $\pm$ 0.0222 & 0.3228 $\pm$ 0.0076 & 0.0220 $\pm$ 0.0035 & \textbf{0.0001} $\pm$ \textbf{4.76e-5} \\
    \bottomrule
  \end{tabular}}
\end{table}

%scaling done
\begin{table}[H]
  \caption{Prediction error (RMSE) and fairness performances on the synthetic dataset. The dataset preserves observation bias $(q_0,q_1)=(0.2, 0.03)$ while exhibiting no population imbalance $(p_0,p_1)=(0.4, 0.4)$.}
  \label{table:synthetic_MF2}
  \centering
  \resizebox{\textwidth}{!}{\begin{tabular}{cccccc}
    \toprule
    Measure & RMSE & $\sf DEE$ & $\sf VAL$ & $\sf UGF$ & $\sf CVS$ \\
    \midrule
    Unfair   & 0.8031 $\pm$ 0.0127 & 0.0783 $\pm$ 0.0053 & 0.2248 $\pm$ 0.0068 & 0.0161 $\pm$ 0.0043 & 0.0073 $\pm$ 0.0040 \\
    Ours (\sf{DEE})   & 0.8240 $\pm$ 0.0074& \textbf{0.0045} $\pm$ \textbf{0.0032} & 0.2427 $\pm$ 0.0089 & \underline{0.0018 $\pm$ 0.0018}& \underline{0.0005 $\pm$ 0.0004}\\
    Ours (\sf{DER})  & 0.8287 $\pm$ 0.0080 & 0.0507 $\pm$ 0.0086 & 0.2495 $\pm$ 0.0063 & 0.0253 $\pm$ 0.0043 & 0.0001 $\pm$ 0.0001\\
    $\sf VAL$-based   & 0.8155 $\pm$ 0.0100 & 0.0779 $\pm$ 0.0042 & \textbf{0.0003} $\pm$ \textbf{9.26e-6} & 0.0087 $\pm$ 0.0014& 0.0082 $\pm$ 0.0013\\
    $\sf UGF$-based   & 0.8150 $\pm$ 0.0111 & 0.0910 $\pm$ 0.0109& 0.2338 $\pm$ 0.0092 & \textbf{0.0008} $\pm$ \textbf{0.0003} & 0.0211 $\pm$ 0.0021 \\
    $\sf CVS$-based  & 0.8151 $\pm$ 0.0115 & 0.0791 $\pm$ 0.0110 & 0.2374 $\pm$ 0.0092 & 0.0234 $\pm$ 0.0014& \textbf{0.0002} $\pm$ \textbf{0.0001} \\
    \bottomrule
  \end{tabular}}
\end{table}

%scaling done
\begin{table}[H]
  \caption{Prediction error (RMSE) and fairness performances on the synthetic dataset. The dataset preserves population imbalance $(p_0,p_1)=(0.4, 0.1)$ while exhibiting no observation bias $(q_0,q_1)=(0.2, 0.2)$.}
  \label{table:synthetic_MF3}
  \centering
  \resizebox{\textwidth}{!}{\begin{tabular}{cccccc}
    \toprule
    Measure & RMSE & $\sf DEE$ & $\sf VAL$ & $\sf UGF$ & $\sf CVS$ \\
    \midrule
    Unfair   & 0.0837 $\pm$ 0.0149 & 0.5859$\pm$ 0.0011 & 0.0727 $\pm$ 0.0009 & 0.0193 $\pm$ 0.0004 & 0.0018 $\pm$ 0.0005 \\
    Ours (\sf{DEE})   & 0.6821 $\pm$ 0.0025 & \textbf{0.0123} $\pm$ \textbf{0.0004} & 0.1865 $\pm$ 0.0096 & \underline{0.0004 $\pm$ 0.0002}& \underline{0.0003 $\pm$ 0.0002}\\
    Ours (\sf{DER}) & 0.6761 $\pm$ 0.0039 & 0.0507 $\pm$ 0.0052 & {0.1885} $\pm$ {0.0100} & 0.0254 $\pm$ 0.0026 & 0.0004 $\pm$ 0.0002\\
    $\sf VAL$-based & 0.3436 $\pm$ 0.0110 & 0.5648 $\pm$ 0.0022 & \textbf{0.0002} $\pm$ \textbf{9.40e-6} & 0.0182 $\pm$ 0.0007 & 0.0018 $\pm$ 0.0006\\
    $\sf UGF$-based   & 0.5640 $\pm$ 0.2033 & 0.4495 $\pm$ 0.1660 & 0.0935 $\pm$ 0.0527 & \textbf{0.0001} $\pm$ \textbf{3.62e-5} & 0.0047 $\pm$ 0.0024 \\
    $\sf CVS$-based   & 0.1277 $\pm$ 0.0107 & 0.5856 $\pm$ 0.0015 & 0.0690 $\pm$ 0.0005 & 0.0188 $\pm$ 0.0006 & \textbf{0.0002} $\pm$ \textbf{0.0001} \\
    \bottomrule
  \end{tabular}}
\end{table}

%scaling done
\begin{table}[H]
  \caption{Prediction error (RMSE) and fairness performances on the synthetic dataset. The dataset preserves population imbalance $(p_0,p_1)=(0.4, 0.2)$ while exhibiting no observation bias $(q_0,q_1)=(0.2, 0.2)$.}
  \label{table:synthetic_MF4}
  \centering
  \resizebox{\textwidth}{!}{\begin{tabular}{cccccc}
    \toprule
    Measure & RMSE & $\sf DEE$ & $\sf VAL$ & $\sf UGF$ & $\sf CVS$ \\
    \midrule
    Unfair   & 0.0600 $\pm$ 0.0102 & 0.3789 $\pm$ 0.0007 & 0.0697 $\pm$ 0.0004& 0.0217 $\pm$ 0.0005 & 0.0007  $\pm$ 0.0003 \\
    Ours (\sf{DEE})   & 0.6125 $\pm$ 0.0038 & \textbf{0.0115} $\pm$ \textbf{0.0009} & 0.0938 $\pm$ 0.0029 & \underline{0.0005 $\pm$ 0.0001}& \underline{0.0003 $\pm$ 0.0002}\\
    Ours (\sf{DER})   & 0.6187 $\pm$ 0.0067 & 0.0622 $\pm$ 0.0015 & 0.0894 $\pm$ 0.0016 & 0.0311 $\pm$ 0.0008 & 0.0006 $\pm$ 0.0002\\
    $\sf VAL$-based   & 0.3451 $\pm$ 0.0109 & 0.3684 $\pm$ 0.0022 & \textbf{0.0002} $\pm$ \textbf{3.03e-6} & 0.0207 $\pm$ 0.0011 & 0.0033 $\pm$ 0.0011\\
    $\sf UGF$-based   & 0.4221 $\pm$ 0.0130 & 0.3713 $\pm$ 0.0041 & 0.0592 $\pm$ 0.0020 & \textbf{0.0002} $\pm$ \textbf{2.38e-5} & 0.0058 $\pm$ 0.0011 \\
    $\sf CVS$-based  & 0.1046 $\pm$ 0.0059 & 0.3790 $\pm$ 0.0007 & 0.0672 $\pm$ 0.0012 & 0.0219 $\pm$ 0.0004 & \textbf{0.0001} $\pm$ \textbf{4.43e-5} \\
    \bottomrule
  \end{tabular}}
\end{table}

%scaling done
\begin{table}[H]
  \caption{Prediction error (RMSE) and fairness performances on the synthetic dataset. The dataset preserves population imbalance $(p_0,p_1)=(0.4, 0.3)$ while exhibiting no observation bias $(q_0,q_1)=(0.2, 0.2)$.}
  \label{table:synthetic_MF5}
  \centering
  \resizebox{\textwidth}{!}{\begin{tabular}{cccccc}
    \toprule
    Measure & RMSE & $\sf DEE$ & $\sf VAL$  & $\sf UGF$  & $\sf CVS$ \\
    \midrule
    Unfair   & 0.0582 $\pm$ 0.0098 & 0.1952 $\pm$ 0.0002 & 0.0678 $\pm$ 0.0006 & 0.0233 $\pm$ 0.0002 & 0.0022 $\pm$ 0.0005 \\
    Ours (\sf{DEE})   & 0.4525 $\pm$ 0.0058 & \textbf{0.0049} $\pm$ \textbf{0.0028} & 0.0767 $\pm$ 0.0023 & \underline{0.0004 $\pm$ 0.0003}& \underline{0.0002 $\pm$ 0.0001}\\
    Ours (\sf{DER})  & 0.4946 $\pm$ 0.0058 & 0.0609 $\pm$ 0.0049 & 0.0827 $\pm$ 0.0021 & 0.0305 $\pm$ 0.0025 & 0.0003 $\pm$ 0.0002\\
    $\sf VAL$-based   & 0.3460 $\pm$ 0.0046 & 0.1863 $\pm$ 0.0018 & \textbf{0.0002} $\pm$ \textbf{6.32e-6} & 0.0246 $\pm$ 0.0010 & 0.0030 $\pm$ 0.0012\\
    $\sf UGF$-based  & 0.4031 $\pm$ 0.0095 & 0.1830 $\pm$ 0.0040 & 0.0683 $\pm$ 0.0024 & \textbf{3.84e-5} $\pm$ \textbf{2.62e-5} & 0.0039 $\pm$ 0.0006 \\
    $\sf CVS$-based  & 0.1077 $\pm$ 0.0178 & 0.1927 $\pm$ 0.0014 & 0.0660 $\pm$ 0.0008 & 0.0228 $\pm$ 0.0011 & \textbf{0.0002} $\pm$ \textbf{2.30e-5} \\
    \bottomrule
  \end{tabular}}
\end{table}

\subsection{Real datasets experiments}
\label{sec: appendix_real}
\begin{itemize}
    \item \emph{MovieLens 1M}: The associated task is to predict the movie rating on a 5-star scale. This dataset contains 6,040 users, 3,900 movies, and 1,000,209 ratings, i.e., rating matrix is 4.26\% full.\footnote{http://www.movielens.org/} We divide user and item groups based on gender and genre, respectively. Action, crime, filme-noir, war are selected as male-preferred genre, whereas children, fantasy, musical, romance are selected as female-preferred genre. We can select male-preferred and female-preferred genres in a variety of ways 
   based on ratings and observations. For various scenarios, the experimental results with similar trends are obtained, so we report the results for one representative scenario. If we assume that the real dataset is generated from the same model as the synthetic dataset, we can estimate the following probabilities. We empirically estimate the interested probabilities w.r.t. population imbalance as: $\widehat{p}_{\mathcal{M}\mathcal{M}'}=0.627,\ \widehat{p}_{\mathcal{M}\mathcal{F}'}=0.517,\  \widehat{p}_{\mathcal{F}\mathcal{M}'}=0.622$, and $\widehat{p}_{\mathcal{F}\mathcal{F}'}=0.595$. Similarly we obtain the estimates for the other probabilities w.r.t. observation bias: $\widehat{q}_{\mathcal{M}\mathcal{M}'}=0.053,\ \widehat{q}_{\mathcal{M}\mathcal{F}'}=0.037,\  \widehat{q}_{\mathcal{F}\mathcal{M}'}=0.037$, and $\widehat{q}_{\mathcal{F}\mathcal{F}'}=0.046$.
    \item \emph{Last FM 360K}: The associated task is to predict whether the user likes the artist or not. This dataset contains 359,347 users, 294,015 artists, and 17,559,530 play counts, i.e., rating matrix is 0.02\% full.\footnote{http://ocelma.net/MusicRecommendationDataset/lastfm-360K.html} The data for play counts is converted to binary rating: $+1$ if counts $>$ average, otherwise $-1$. We divide user and item groups based on gender and genre, respectively. Since this dataset only contains gender information, we use Last.fm API\footnote{http://www.last.fm/api} to collect the genre of corresponding artist's music; the tag was associated with 5,706 artists. 
    We also randomly select 5000 male and 5000 female users. Among 10 genres, we choose hip-hop and musical for male and female preferred genres, respectively. The final rating matrix of 10,000 users and 5,706 artists is 0.55\% full. From the real data, we obtain empirical estimates for the interested probabilities w.r.t. population imbalance: $\widehat{p}_{\mathcal{M}\mathcal{M}'}=0.548,\ \widehat{p}_{\mathcal{M}\mathcal{F}'}=0.421,\  \widehat{p}_{\mathcal{F}\mathcal{M}'}=0.438$ and $\widehat{p}_{\mathcal{F}\mathcal{F}'}=0.529$. Similarly we obtain the estimates for the other probabilities w.r.t. observation bias: $\widehat{q}_{\mathcal{M}\mathcal{M}'}=0.0054,\ \widehat{q}_{\mathcal{M}\mathcal{F}'}=0.0011,\  \widehat{q}_{\mathcal{F}\mathcal{M}'}=0.0036$ and $\widehat{q}_{\mathcal{F}\mathcal{F}'}=0.0038$.
\end{itemize}

We randomly split the real datasets into 90\% train set and 10\% test set. In case of MovieLens data, the rating is five-star based, so we set the threshold $\tau=3$, i.e., $\widetilde{Y}={\bf 1} \{ \widehat{Y} \geq 3 \}$. On the other hand, for LastFM dataset, we set $\tau=0$ as $M_{ij} \in \{+1, -1\}$. We run experiments employing both matrix factorization (MF) based and autoencoder (AE) based techniques.  We set the rank as 512 for MF-based algorithm as was found by hyperparameter search. The structure of the employed autoencoder~\citep{sedhain2015} is as follows: (i) encoder has two linear layers: 512 nodes with ReLU actiavation and 512 nodes with dropout layer ($\textrm{rate}=0.7$) and ReLU activation; (ii) decoder has one layer with 512 nodes. For MovieLens 1M data (five-star ratings), we apply the clipping to the decoder output to fit into the range. For LastFM 360K data (binary rating: $+1$ and $-1$), we apply tanh activation. Hyperparameters for KDE-based algorithm are set to $(\delta,h)=(0.01, 0.01)$ and $\lambda=0.9$. We use Adam optimizer for 1,000 iterations using full gradient, and the learning rate is set to 1e-3. Since the main paper contains mostly MF-based experiments, here we only present the performances of \emph{autoencoder} based algorithm on both real datasets.  
% after the hyperparameter search in \{1e-2, 1e-3, 1e-4, 1e-5, 1e-6\}. 
 %We repeat the experiment five times for each settings and print out the result in average and standard deviation. 

%scaling done.
\begin{table}[H]
  \caption{Prediction error (RMSE) and fairness performances of the \emph{autoencoder} based algorithm on MovieLens 1M dataset. We observe the same performance trends as those in Table~\ref{table:synthetic_MF1}.}
  \label{table:movielens_AE}
  \centering
  \resizebox{\textwidth}{!}{\begin{tabular}{cccccc}
    \toprule
    Measure & RMSE & $\sf DEE$  & $\sf VAL$ & $\sf UGF$ & $\sf CVS$\\
    \midrule
    Unfair   & 0.8369 $\pm$ 0.0012     & 0.2477 $\pm$ 0.0175 & 0.3412 $\pm$ 0.0031    & 0.0419 $\pm$ 0.0042   & 0.1158 $\pm$ 0.0025 \\
    Ours (\sf{DEE})   & 0.8437 $\pm$ 0.0042     & \textbf{0.0120} $\pm$ \textbf{0.0028} & 0.3338 $\pm$ 0.0037    & \underline{0.0039 $\pm$ 0.0022}  & \underline{0.0042 $\pm$ 0.0010} \\
    Ours (\sf{DER})  & 0.8411 $\pm$ 0.0027     & 0.0285 $\pm$ 0.0084 & {0.3395} $\pm$ {0.0048}  & 0.0144 $\pm$ 0.0046 & 0.0061 $\pm$ 0.0023 \\
    $\sf VAL$-based  & 0.8433 $\pm$ 0.0022     & 0.2138 $\pm$ 0.0363 & \textbf{0.2128} $\pm$ \textbf{0.0143}  & 0.0299 $\pm$ 0.0172 & 0.0918 $\pm$ 0.0070 \\
    $\sf UGF$-based  & 0.8491 $\pm$ 0.0056     & 0.1934 $\pm$ 0.0109  & 0.3391 $\pm$ 0.0040  & \textbf{0.0011} $\pm$ \textbf{0.0006} & 0.0982 $\pm$ 0.0055 \\
    $\sf CVS$-based  & 0.8495 $\pm$ 0.0050     & 0.0808 $\pm$ 0.0225 & 0.3424 $\pm$ 0.0069  & 0.0343 $\pm$ 0.0105 & \textbf{0.0023} $\pm$ \textbf{0.0010} \\
    \bottomrule
  \end{tabular}}
\end{table}

%scaling done.
\begin{table}[H]
  \caption{Prediction error (RMSE) and fairness performances of the \emph{autoencoder} based algorithm on Last FM 360K dataset.}
  \label{table:LastFM_AE}
  \centering
  \resizebox{\textwidth}{!}{\begin{tabular}{cccccc}
    \toprule
    Measure & RMSE & $\sf DEE$  & $\sf VAL$ & $\sf UGF$ & $\sf CVS$ \\
    \midrule
    Unfair   & 0.6534 $\pm$ 0.0024  & 0.1003 $\pm$ 0.0172   & 0.2253 $\pm$ 0.0031  & 0.0501 $\pm$ 0.0086   & 0.0006 $\pm$ 0.0002  \\
     Ours (\sf{DEE})   & 0.6649 $\pm$ 0.0212     & \textbf{0.0024} $\pm$ \textbf{0.0005}  & 0.2213 $\pm$ 0.0133    & \underline{0.0012 $\pm$ 0.0008}  & \underline{0.0007 $\pm$ 0.0006} \\
     Ours (\sf{DER}) & 0.6501 $\pm$ 0.0004     & 0.0878 $\pm$ 0.0015 
     & 0.2204 $\pm$ 0.0004 & 0.0439 $\pm$ 0.0008 & 0.0008 $\pm$ 0.0001\\
     $\sf VAL$-based & 0.6828 $\pm$ 0.0202     & 0.0571 $\pm$ 0.0054 
     & \textbf{0.1915} $\pm$ \textbf{0.0038}  & 0.0288 $\pm$ 0.0029 & 0.0063 $\pm$ 0.0029\\
    $\sf UGF$-based   & 0.6861 $\pm$ 0.0310     & 0.0485 $\pm$ 0.0025    & 0.2098 $\pm$ 0.0143  & \textbf{0.0001} $\pm$ \textbf{5.00e-5} & 0.0021 $\pm$ 0.0016\\
    $\sf CVS$-based & 0.6685 $\pm$ 0.0079     & 0.0960 $\pm$ 0.0096    & 0.2421 $\pm$ 0.0105  & 0.0480 $\pm$ 0.0047 & \textbf{0.0002} $\pm$ \textbf{3.24e-5}\\
    \bottomrule
  \end{tabular}}
\end{table}

\begin{figure}[h!]
    \centering  
    \includegraphics[height=4.5cm]{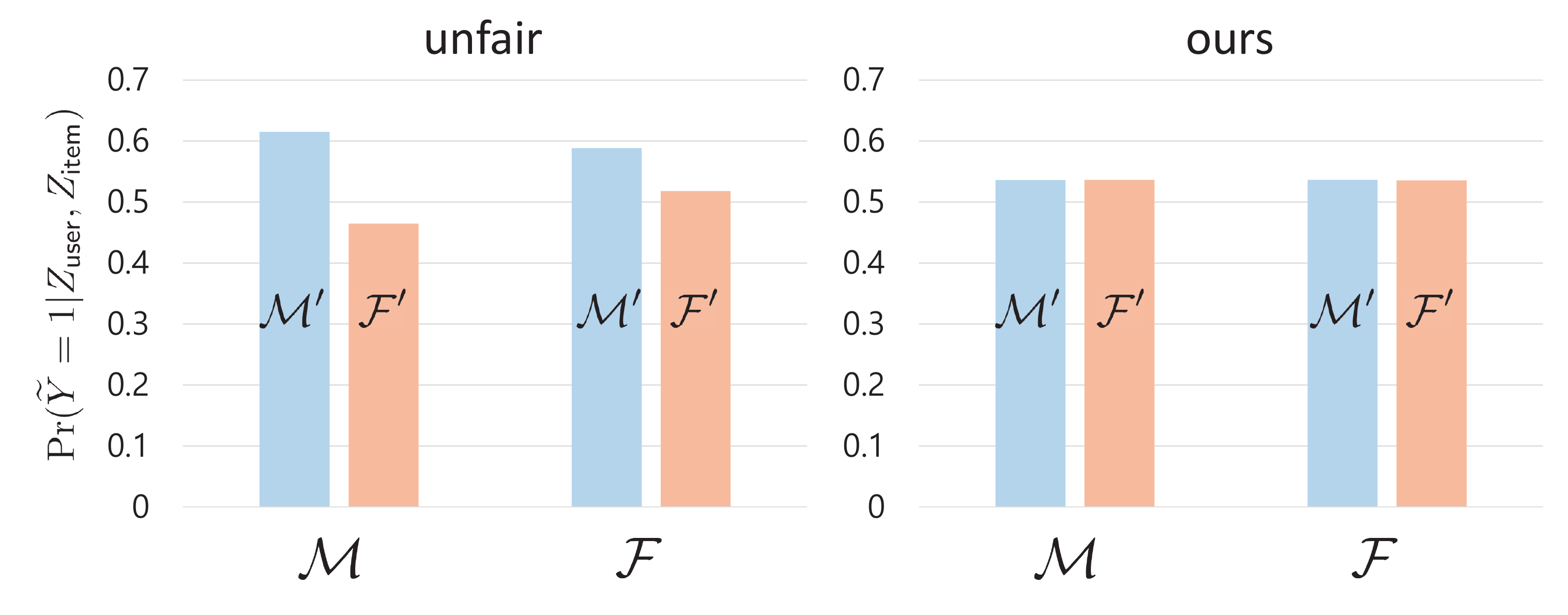}
    \caption{{Predicted preference rate of item groups for every user group $\Pr(\tilde{Y}=1|Z_{\sf user}, Z_{\sf item})$ of the matrix factorization based algorithm on \emph{MovieLens 1M dataset}.}}
    \label{fig:5}
\end{figure}

\begin{figure}[h!]
    \centering  
    \includegraphics[height=4.5cm]{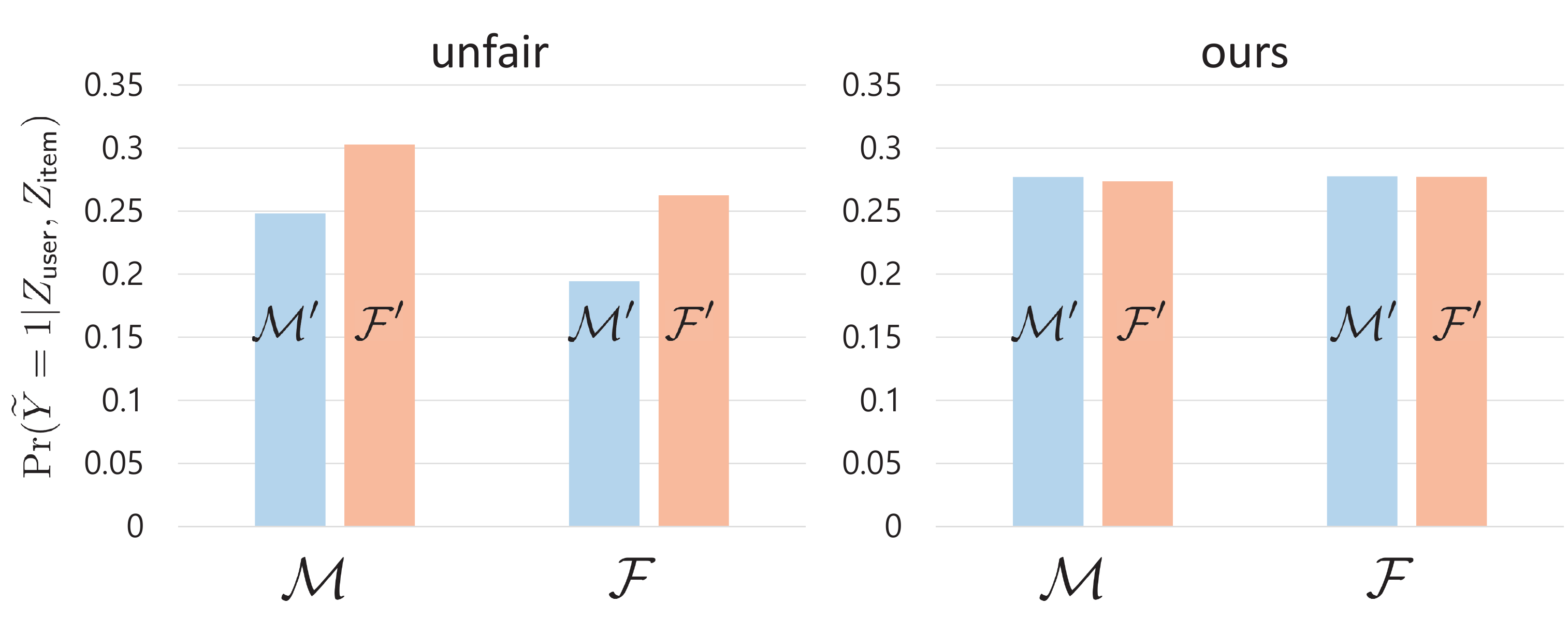}
    \caption{{Predicted preference rate of item groups for every user group $\Pr(\tilde{Y}=1|Z_{\sf user}, Z_{\sf item})$ of the matrix factorization based algorithm on \emph{Last FM 360K dataset}.}}
    \label{fig:6}
\end{figure}

\begin{figure}[h!]
    \centering  
    \includegraphics[height=4.5cm]{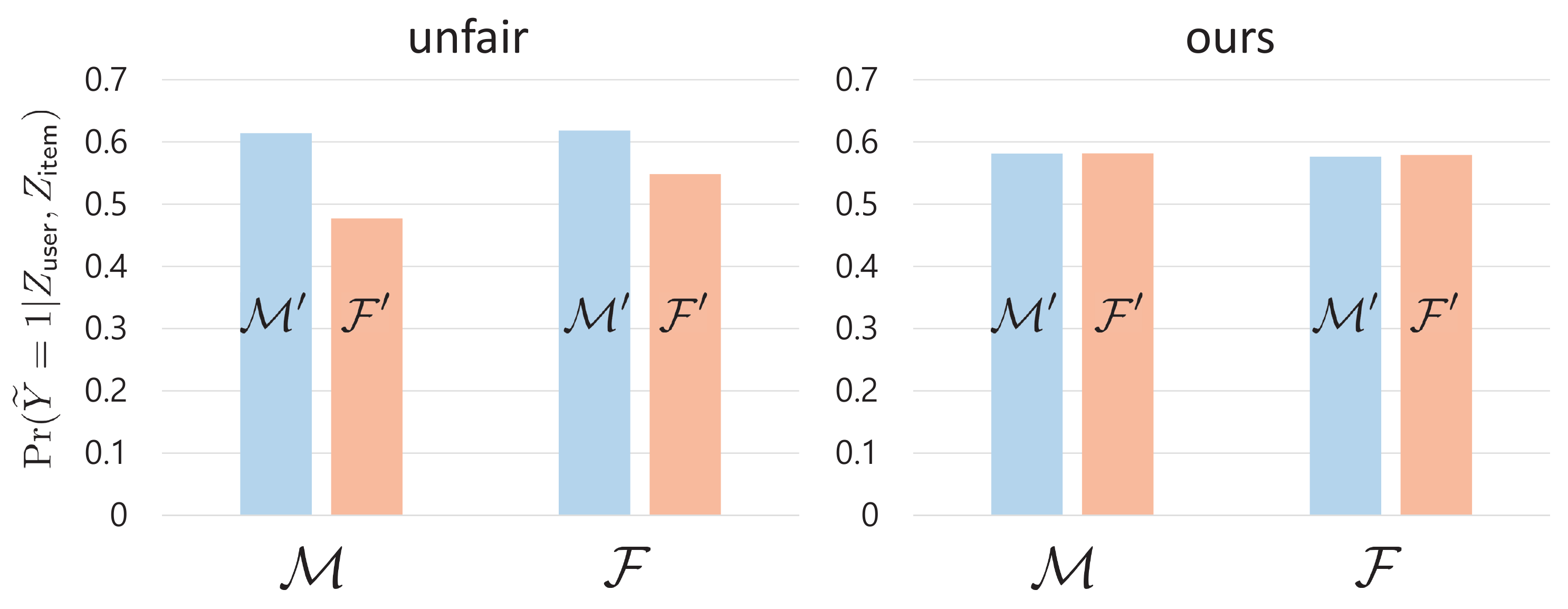}
    \caption{{Predicted preference rate of item groups for every user group $\Pr(\tilde{Y}=1|Z_{\sf user}, Z_{\sf item})$ of the \emph{autoencoder} based algorithm on \emph{MovieLens 1M dataset}.}}
    \label{fig:7}
\end{figure}

\begin{figure}[h!]
    \centering  
    \includegraphics[height=4.5cm]{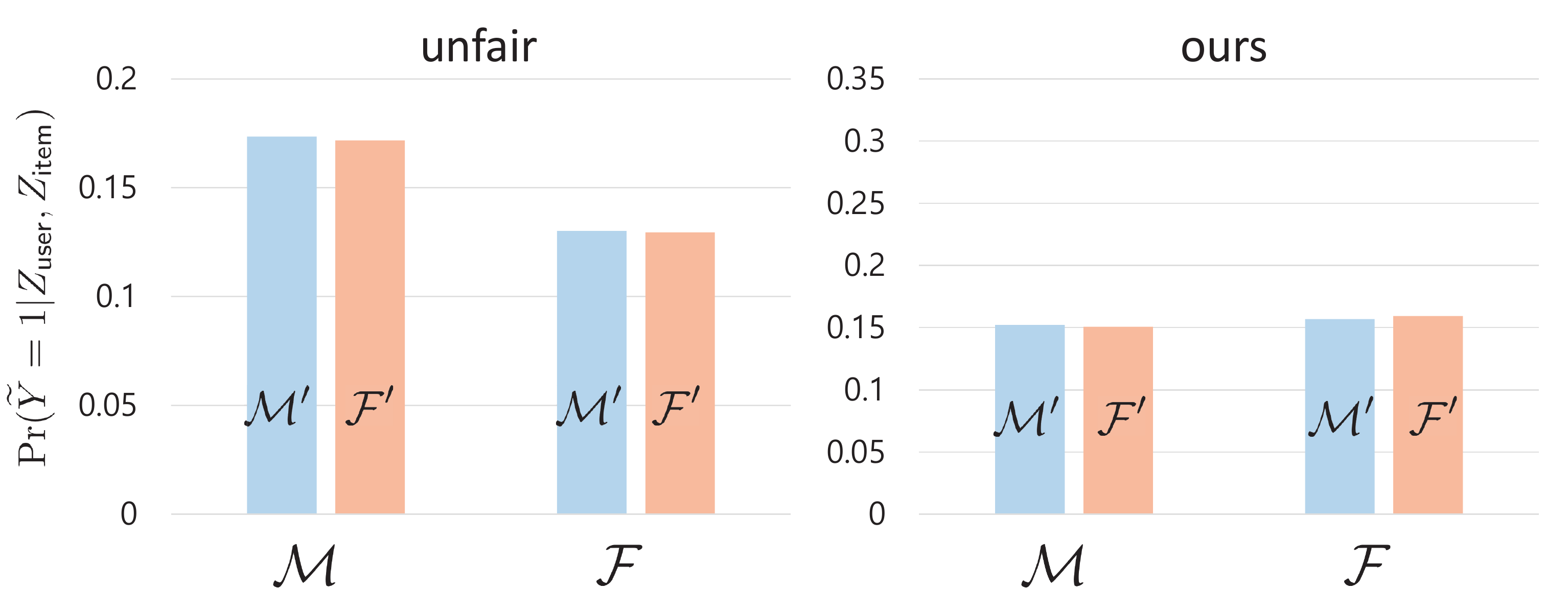}
    \caption{{Predicted preference rate of item groups for every user group $\Pr(\tilde{Y}=1|Z_{\sf user}, Z_{\sf item})$ of the \emph{autoencoder} based algorithm on \emph{Last FM 360K dataset}.}}
    \label{fig:8}
\end{figure}

\subsection{Complexity analysis}
\label{sec:appendix_running_times}
We do complexity analysis of ours in light of other baselines. For comparison, we consider the running time measured under Pytorch on Xeon Silver 4210R CPU and TITAN RTX GPU. Table~\ref{table:running_time} presents the running times of matrix factorization based algorithms on the synthetic and real datasets. While our approach provides better fairness performance w.r.t. $\sf DEE$ (as in the above tables), it comes at a cost of an increased complexity, around twice relative to the $\sf CVS$-based algorithm.

\begin{table}[H]
  \caption{The running time (in seconds) of our algorithm and baselines on the synthetic and two real datasets: MovieLens 1M and LastFM 360K.}
  \label{table:running_time}
  \centering
  {\begin{tabular}{cccccc}
    \toprule
    Dataset & Synthetic & MovieLens 1M & LastFM 360K\\
    \midrule
   Unfair   & 2.15 & 6.72 & 16.27  \\
    Ours (\sf{DEE})   & 13.23 & 86.82 & 192.02  \\
      $\sf VAL$-based~\citep{yao2017}   & 5.83 & 201.08 & 477.14  \\
       $\sf UGF$-based~\citep{li2021}    & 7.05 & 60.14 & 136.69  \\
       $\sf CVS$-based~\citep{kamishima2017}  & 7.16 & 47.89 & 104.42  \\
    \bottomrule
  \end{tabular}}
\end{table}

\subsection{Extension to the fairness notion: $\widetilde{Y}\perp Z_{\sf user}, Z_{\sf item}|Y$}
\label{sec: extended fairness notion}
Similar to \textsf{DEE} in equation~\ref{def:DEE}, we quantify the notion via: 
\begin{equation}\label{DEO}
    \sum_{y\in\{0,1\}}\sum_{z_1 \in\mathcal{Z}_{\sf user}}\sum_{z_2 \in\mathcal{Z}_{\sf item}}
    \Big \vert \mathbb{P} (\widetilde{Y}=1|Y=y) -\mathbb{P}(\widetilde{Y}=1|Y=y, Z_{\sf item} = z_2,Z_{\sf user}=z_1) \Big \vert,
\end{equation}
for arbitrary alphabet sizes $|\mathcal{Z}_{\sf user}|$ and $|\mathcal{Z}_{\sf item}|$. 

Using the KDE approach, similarly we can obtain:
\begin{align*}
  \widehat{ \mathbb{P} } (\widetilde{Y} = 1|Y=y) 
  =\int_{\tau}^{\infty}\widehat{f}_{\widehat{Y}|Y}(\hat{y}|y)d\hat{y}=\frac{1}{|I_y| }\sum_{(i, j)\in I_y}
  F_k\left({\frac{\tau-\widehat{M}_{ij} }{h}}\right),
\end{align*}
where $F_k(\hat{y}):=\int_{\hat{y}}^{\infty}f_k(t)dt$; $Y_{ij}:= \mathbf{1} \{ M_{ij} \geq \tau \}$; and $I_y:=\{(i,j): (i,j)\in\Omega, Y_{ij}=y\}$. We can then compute the gradien w.r.t. $w$ as: 
\begin{align}
\label{eq:DP_grad_ext}
    \nabla_w\widehat{\mathbb{P}}(\widetilde{Y}=1|Y=y)=
    \frac{1}{|I_y|h}\sum_{(i,j)\in I_{y}}f_k\left({\frac{\tau-\widehat{M}_{ij}}{h}}\right)\cdot\nabla_w{\widehat{M}_{ij}}.
\end{align}
We can then enjoy a family of gradient-based optimizers~\citep{Geron, ref20_adam}.
We provide experimental results for the extension on MovieLens 1M~\citep{data_movielens} real dataset. We run experiments employing both matrix factorization (MF) based and autoencoder (AE) based techniques. We demonstrate that the framework based on the extended notion can indeed mitigate such unfairness while exhibiting a minor degradation of recommendation accuracy. The results are listed in Table~\ref{table:movielens_MF_extension} and~\ref{table:movielens_AE_extension}.

\begin{table}[H]
  \caption{Prediction error (RMSE) and fairness performances of the matrix factorization based algorithm on \emph{MovieLens 1M dataset}. The boldface indicates the best result and the underline denotes the second best. The approach based on the extended fairness notion (conditioning on $Y$), enjoys the best fairness performance for the measure focused therein.}
  \label{table:movielens_MF_extension}
  \centering
  \resizebox{\textwidth}{!}{\begin{tabular}{ccccccc}
    \toprule
    Measure & RMSE & $\sf DEE$ & Conditioning on $Y$~(\ref{DEO}) & $\sf VAL$ & $\sf UGF$ & $\sf CVS$ \\
    \midrule
    Unfair   & 0.8541 $\pm$ 0.0033     & 0.2447 $\pm$ 0.0134 & 0.3494 $\pm$ 0.0071 & 0.3227 $\pm$ 0.0031    & 0.0058 $\pm$ 0.0042   & 0.1291 $\pm$ 0.0079  \\
    Ours (\sf{DEE})   & 0.8641 $\pm$ 0.0047     & \textbf{0.0014} $\pm$ \textbf{0.0008}  & 0.3005 $\pm$ 0.0048 & 0.2941 $\pm$ 0.0024    & \underline{0.0018 $\pm$ 0.0016}  & \underline{0.0007 $\pm$ 0.0004} \\
    Conditioning on $Y$ & 0.8576 $\pm$ 0.0011     & 0.1451 $\pm$ 0.0079 & \textbf{0.0283 $\pm$ 0.0057} & 0.3230 $\pm$ 0.0026    & 0.0052 $\pm$ 0.0029  & 0.0668 $\pm$ 0.0051 \\
    $\sf VAL$-based   & 0.8529 $\pm$ 0.0011    & 0.3659 $\pm$ 0.0033  & 0.4679 $\pm$ 0.0098  & \textbf{0.0942} $\pm$ \textbf{0.0016}  & 0.0261 $\pm$ 0.0020 & 0.1388 $\pm$ 0.0030\\
    $\sf UGF$-based   & 0.8550 $\pm$ 0.0015     & 0.2492 $\pm$ 0.0100  & 0.3657 $\pm$ 0.0084   & 0.3285 $\pm$ 0.0051  & \textbf{0.0001} $\pm$ \textbf{0.0001} & 0.1355 $\pm$ 0.0038\\
    $\sf CVS$-based   & 0.8549 $\pm$ 0.0018     & 0.0721 $\pm$ 0.0069  & 0.3260 $\pm$ 0.0135   & 0.3319 $\pm$ 0.0046  & 0.0065 $\pm$ 0.0042 & \textbf{0.0002} $\pm$ \textbf{3.45e-5}\\
    \bottomrule
  \end{tabular}}
\end{table}

%scaling done.
\begin{table}[H]
  \caption{Prediction error (RMSE) and fairness performances of the \emph{autoencoder} based algorithm on MovieLens 1M dataset.}
  \label{table:movielens_AE_extension}
  \centering
  \resizebox{\textwidth}{!}{\begin{tabular}{ccccccc}
    \toprule
    Measure & RMSE & $\sf DEE$ & Conditioning on $Y$~(\ref{DEO}) & $\sf VAL$ & $\sf UGF$ & $\sf CVS$\\
    \midrule
    Unfair   & 0.8369 $\pm$ 0.0012     & 0.2477 $\pm$ 0.0175 & 0.3557 $\pm$ 0.0086 & 0.3412 $\pm$ 0.0031    & 0.0419 $\pm$ 0.0042   & 0.1158 $\pm$ 0.0025 \\
   Ours   & 0.8437 $\pm$ 0.0042     & \textbf{0.0120} $\pm$ \textbf{0.0028} & 0.2260 $\pm$ 0.0203 & 0.3338 $\pm$ 0.0037    & \underline{0.0039 $\pm$ 0.0022}  & \underline{0.0042 $\pm$ 0.0010} \\
    Conditioning on $Y$ & 0.8467 $\pm$ 0.0012 & 0.1157 $\pm$  0.0203 & \textbf{0.0505 $\pm$ 0.0264} & 0.3395 $\pm$ 0.0039    & 0.0131 $\pm$ 0.0139  & 0.0551 $\pm$ 0.0046 \\
    $\sf VAL$-based  & 0.8433 $\pm$ 0.0022     & 0.2138 $\pm$ 0.0363 & 0.3372 $\pm$ 0.0117  & \textbf{0.2128} $\pm$ \textbf{0.0143}  & 0.0299 $\pm$ 0.0172 & 0.0918 $\pm$ 0.0070 \\
    $\sf UGF$-based  & 0.8491 $\pm$ 0.0056     & 0.1934 $\pm$ 0.0109 & 0.3285 $\pm$ 0.0178  & 0.3391 $\pm$ 0.0040  & \textbf{0.0011} $\pm$ \textbf{0.0006} & 0.0982 $\pm$ 0.0055 \\
    $\sf CVS$-based  & 0.8495 $\pm$ 0.0050     & 0.0808 $\pm$ 0.0225 & 0.2408 $\pm$ 0.0165 & 0.3424 $\pm$ 0.0069  & 0.0343 $\pm$ 0.0105 & \textbf{0.0023} $\pm$ \textbf{0.0010} \\
    \bottomrule
  \end{tabular}}
\end{table}

\subsection{Extension to top-$K$ recommendation}
\label{sec: appendix_recommendation}

\begin{figure}[h]
\begin{center}
    \includegraphics[height=5cm]{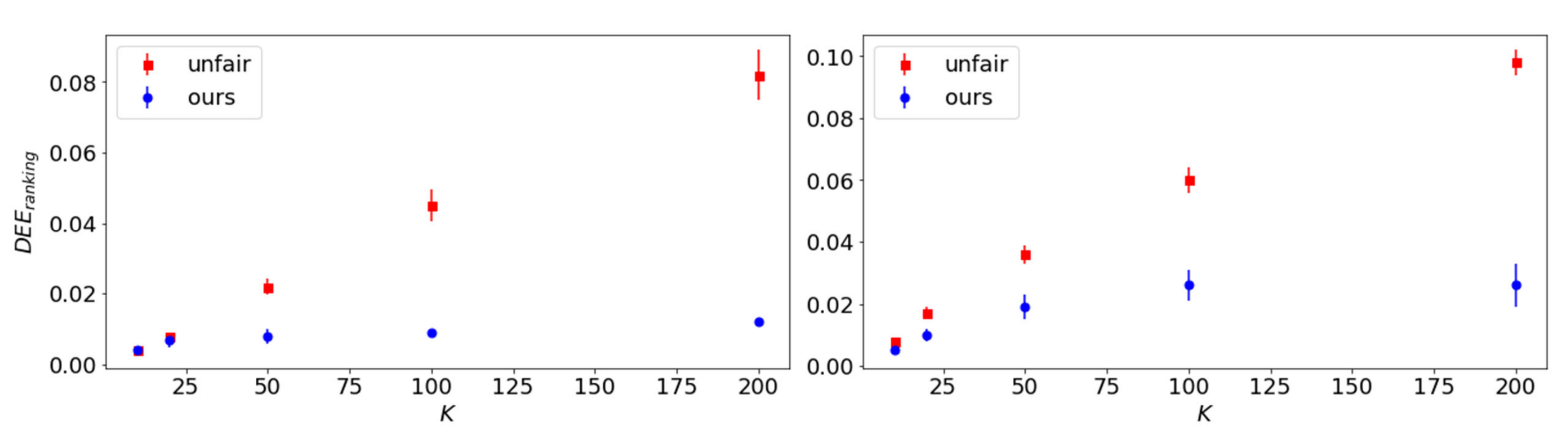}
    \caption{(Left) $\sf{DEE}_{\sf ranking}$ performances of the matrix factorization based algorithms on MovieLens 1M dataset; (Right) the performances of the autoencoder based algorithms on the same dataset. We observe that ours can effectively mitigate unfairness in the context of top-$K$ recommendation.}
    \label{fig:9}
\end{center}
\end{figure}

In this section, we quantify the fairness performance in the context of top-$K$ recommendation. To this end, we generate the end ranked list for each user based on estimated ratings. We then define an indicator function $R$ which returns 1 when the item of interest belongs to, say top-$K$ item set (0 otherwise). Like the equal experience, the notion $R\perp (Z_{\sf user}, Z_{\sf item})$ serves a proper role in this context.  
Similar to \textsf{DEE} in equation~\eqref{def:DEE}, we quantify the notion via:  
\begin{equation}\label{DEE_ranking}
    \textsf{DEE}_{\sf ranking}:=\sum_{z_1 \in\mathcal{Z}_{\sf user}}\sum_{z_2 \in\mathcal{Z}_{\sf item}}
    \Big \vert \mathbb{P} (R=1) -\mathbb{P}(R=1|Z_{\sf item} = z_2,Z_{\sf user}=z_1) \Big \vert.
\end{equation}
In Fig.~\ref{fig:9}, we plot $\textsf{DEE}_{\sf ranking}$ performances as a function of $K\in\{10, 20, 50, 100, 200\}$. We compare the performances on MovieLens 1M dataset for two algorithms: an unfair algorithm (no fairness constraint) and ours (based on DEE in~\eqref{def:DEE}). We observe that ours which builds upon $\sf DEE$ can also effectively mitigate unfairness in the context of top-$K$ recommendation while the unfair algorithm does not work well under the fairness measure as $K$ increases. Here we remark that ${\sf DEE}_{\sf ranking}$ is employed only for the purpose of fairness evaluation in the top-$K$ recommendation setting. In this experiment, we set the threshold which is a design parameter for our algorithm, $\tau=4$, i.e., $\widetilde{Y}=\mathbf{1}\{\widehat{Y}\geq 4\}$ in~\eqref{def:DEE}. Each point and bar represent the average and standard deviation over five trials with different random seeds, respectively.

%%%%%%%%%%%%%%%%%%%%%%%%%%%%%%%%%%%%%%%%%%%%%%%%%%%%%%%%%%%%%%%%%%%%%%%%%%%%%%%
%%%%%%%%%%%%%%%%%%%%%%%%%%%%%%%%%%%%%%%%%%%%%%%%%%%%%%%%%%%%%%%%%%%%%%%%%%%%%%%

\end{document}